\pdfoutput=1
\documentclass{article}

\usepackage[english]{babel}

\usepackage[letterpaper,top=2cm,bottom=2cm,left=3cm,right=3cm,marginparwidth=1.75cm]{geometry}

\usepackage[parfill]{parskip}
\newlength{\defbaselineskip}
\usepackage[symbol]{footmisc}

\usepackage{svg}
\usepackage{amsmath}
\usepackage{graphicx}
\graphicspath{ {figures/} }
\usepackage[colorlinks=true, allcolors=blue]{hyperref}
\usepackage{authblk}

\usepackage[utf8]{inputenc} 
\usepackage[T1]{fontenc}    
\usepackage{url}            
\usepackage{booktabs}       
\usepackage{amsfonts}       
\usepackage{nicefrac}       
\usepackage{microtype}      
\usepackage{amssymb}
\usepackage{subfigure}
\usepackage{enumitem}
\usepackage{amsthm}     
\usepackage{cleveref}   
\usepackage{wrapfig}
\usepackage{subcaption}
\usepackage{multirow}
\usepackage{algorithm}
\usepackage[noend]{algpseudocode}
\usepackage{tabulary}
\usepackage{csquotes}
\usepackage{bbm}
\usepackage{tabularx}

\usepackage[
  backend=biber,
  citestyle=authoryear-comp,  
  style=authoryear,           
  sorting=nyt,
  maxcitenames=1,             
  mincitenames=1,
  maxbibnames=99,             
  minbibnames=1,
  uniquename=false,           
  uniquelist=false            
]{biblatex}
\allowdisplaybreaks
\usepackage[symbol]{footmisc}
\usepackage{soul}
\renewcommand{\thefootnote}{\fnsymbol{footnote}}

\title{Scaling Fine-Grained MoE Beyond 50B Parameters:\\Empirical Evaluation and Practical Insights}

\author{Jakub Krajewski$^{2}$\footnote{Work done during internship at NVIDIA.  Correspondence to: Jakub Krajewski \texttt{gim.jakubk@gmail.com}} }
\author{Marcin Chochowski$^{1}$}
\author{Daniel Korzekwa$^{1}$}
\affil{$^{1}$NVIDIA \ \ \ $^{2}$IDEAS NCBR, University of Warsaw  \hspace{0.25cm}}

\date{}

\newcommand{\citep}{\parencite}
\newcommand{\citet}{\textcite}

\begin{document}
\maketitle

\definecolor{PlotlyBlue}{HTML}{636EFA}
\definecolor{PlotlyRed}{HTML}{EF553B}

\begin{abstract}
\renewcommand{\thefootnote}{\arabic{footnote}}
\makeatletter
\renewcommand{\@makefntext}[1]{%
    \parindent 1em%
    \noindent
    \hb@xt@1.8em{\hss\@makefnmark}#1}
\makeatother

Mixture of Experts (MoE) architectures have emerged as pivotal for scaling Large Language Models (LLMs) efficiently. Fine-grained MoE approaches - utilizing more numerous, smaller experts - have demonstrated potential in improving model convergence and quality. This work proposes a set of training recipes and provides a comprehensive empirical evaluation of fine-grained MoE, directly comparing its scaling properties against standard MoE configurations for models with up to 56B total (17B active) parameters. We investigate convergence speed, model performance on downstream benchmarks, and practical training considerations across various setups. Overall, at the largest scale we show that fine-grained MoE achieves better validation loss and higher accuracy across a set of downstream benchmarks.
This study offers empirical grounding and practical insights for leveraging fine-grained MoE in the development of future large-scale models.
 
\end{abstract}

\section{Introduction}

Extraordinary capabilities shown in recent years by Large Language Models (LLMs) \citep{openai2024gpt4technicalreport, geminiteam2024geminifamilyhighlycapable} come with steep resource requirements, prompting the search for more cost-effective training methods \citep{faiz2024llmcarbon, touvron2023llama}. Among such approaches, Mixture of Experts (MoE) \citep{shazeer2017outrageously} has emerged as a particularly successful technique, activating only a fraction of the total parameters for each input token. Building upon the standard MoE architecture, fine-grained Mixture of Experts \citep{dai2024deepseekmoe, krajewski2024scalinglawsfinegrainedmixture} leverages a larger number of smaller experts and routes tokens to multiple experts at once, maintaining computational efficiency while often yielding faster convergence.

\vspace{-0.1cm}

This work presents a set of training recipes to effectively scale fine-grained MoE up to 56B total model parameters.
We conduct an empirical evaluation comparing this approach against two standard MoE variants: Top-1 (Switch \citep{fedus2022switch}) and Top-2 (GlaM \citep{du2022glam}, Mixtral \citep{jiang2024mixtral}). All of the experiments use the same controlled setup, including dataset and evaluation protocol, allowing for a fair comparison. We consider two model sizes (11B and 56B total parameters) and multiple training durations, allowing to compare the results in different setups. We measure both the pretraining loss convergence speed and downstream accuracy on a set of benchmarks.

Our key contributions are summarized as follows:
\vspace{-0.1cm}
\begin{itemize}
    \item We compare the convergence of the models across different model and dataset sizes, scaling up to 56B total (17B active) parameters. On the largest scale, we observe gains from increasing granularity in both baseline configurations (Fig. \ref{fig:loss_56b}).
    \item Beyond perplexity-based comparison, we evaluate the models on a set of benchmarks. We demonstrate that the loss improvements transfer to downstream accuracy (Table~\ref{tab:10b_res} \& Table~\ref{tab:eval_largest}).
    \item We analyze and discuss hyperparameter choices and training specifics of fine-grained MoE models (Sec.~\ref{sec:analysis_moe}). Specifically, we evaluate the expert load imbalance, present the evolution of router logits, and highlight the importance of the order of softmax and Top-$k$ normalization in the router.
\end{itemize}

\section{Related Work}

\paragraph{Mixture of Experts.} The Mixture of Experts layer has been originally introduced by \citep{shazeer2017outrageously}, and further developed in a series of works, including Switch \citep{fedus2022switch}, GShard \citep{lepikhin2020gshard}, GLaM \citep{du2022glam}, ST-MoE \citep{zoph2022stmoedesigningstabletransferable}, and more recently OpenMoE \citep{xue2024openmoeearlyeffortopen}, Mixtral \citep{jiang2024mixtral} and OLMoE \citep{muennighoff2024olmoeopenmixtureofexpertslanguage}.

\vspace{0.3cm}

In the context of Transformers, MoE is constructed by replacing the Feed-Forward component with a set of \textit{experts}. Each expert is typically a subnetwork of the same shape as the corresponding Feed-Forward layer. Each input $x$ is routed to a subset of experts. The output $y$ of the layer can be defined as 
\begin{align*}
    y = \Sigma_{i\in \tau} R_i(x)E_i(x),
\end{align*}
where $\tau$ is the set of selected selected Top-$k$ experts for this input and $R$ is the router network defining the expert scores for the given input. This router is a simple linear layer, followed by softmax normalization and Top-$k$ choice.
\vspace{0.3cm}

\paragraph{Fine-Grained MoE.} \citep{dai2024deepseekmoe, krajewski2024scalinglawsfinegrainedmixture} propose to relax the assumption that each MoE expert is the same size as the standard Feed-Forward layer. Instead, we can use more, smaller experts, increasing the flexibility of mapping tokens to experts. 
\vspace{0.3cm}

For the Transformer with a hidden Feed-Forward size $d_\text{ff},$ in the standard MoE layer we will choose $k$ experts for each token, with the expert hidden size $d_\text{expert}=d_\text{ff},$ and total number of experts $N_E.$ Then, for fine-grained MoE with granularity $G,$ we increase the total number of experts to $G\cdot N_E$, decrease the expert hidden size to $d_\text{expert}=d_\text{ff} / G,$ and choose $G\cdot k$ experts for each token. This way, the non-router FLOPs and parameters remain unchanged.

\vspace{0.3cm}

Using granularity $G>1$ is the optimal choice from the scaling laws perspective \citep{krajewski2024scalinglawsfinegrainedmixture}. Furthermore, architectures using fine-grained MoE have shown superior performance \citep{deepseekai2024deepseekv2strongeconomicalefficient, deepseekai2024deepseekv3}. However, they often also use other architecture advancements, potentially contributing to the overall efficiency improvement. 
To better understand the scaling advantage of fine-grained MoE, in this report we present the details of our ablations and pre-training experiments on up to 56B of total parameters. 
Additionally, we show downstream evaluations, a case which was not considered in \citep{krajewski2024scalinglawsfinegrainedmixture}, who only focused on modeling perplexity.

\vspace{0.3cm}

\section{Scaling and Evaluation of Fine-Grained MoE}

In this section, we evaluate and compare standard and fine-grained MoE variants through several experiments. 
We start by outlining our experimental setup (Sec.~\ref{sub:exp_setup}). 
We then present the results, beginning with 11B total parameter models (Sec.~\ref{sec:11b}), followed by an analysis of how different training durations affect performance (Sec.~\ref{sec:longer}). 
Finally, we scale the models to 56B total parameters in Sec.~\ref{sec:50b}.

\subsection{Experimental setup} \label{sub:exp_setup}

\paragraph{Training framework and hardware.} The models presented in this report have been trained on a cluster featuring NVIDIA H100 GPUs. We use the openly available Megatron-LM framework \citep{shoeybi2020megatronlmtrainingmultibillionparameter}, employing Pipeline, Expert and Tensor Parallelism. Megatron-LM, in combination with NVIDIA GPUs, enables state-of-the-art efficiency for training both dense and MoE LLMs, achieving >46\% Model FLOPs Utilization (MFU) in training of standard MoE models \citep{vavre2024llama3meetsmoe}.

\paragraph{Model architecture.} We use a standard decoder-only Transformer \citep{vaswani2023attention} architecture, replacing each feed-forward block with a Mixture-of-Experts layer. The models use SwiGLU \citep{shazeer2020gluvariantsimprovetransformer} activation and rotary position embeddings (RoPE) with the rotary percentage set to 0.5. We use the tokenizer from \citep{parmar2024nemotron415btechnicalreport}, with a vocabulary size of 256{,}000.

\newpage

We compare four MoE variants: two common baselines and their fine-grained counterparts. In each pair, the standard and fine-grained models are matched in parameters and compute (excluding router):
\begin{itemize}
    \item A Switch-like MoE with 8 experts and Top-1 routing (denoted as 1$\times$FLOPs-G1), and a fine-grained version with 64 experts and Top-8 routing (1$\times$FLOPs-G8).
    \item A Mixtral-like MoE with 8 experts and Top-2 routing \citep{jiang2024mixtral} (2$\times$FLOPs-G1), and its fine-grained counterpart with 64 experts and Top-16 routing (2$\times$FLOPs-G8).
\end{itemize}

 These fine-grained models correspond to granularity 8 as defined in \citep{krajewski2024scalinglawsfinegrainedmixture}. We find setting $G=8$ sufficient to compare standard and fine-grained MoEs. Higher $G$ would likely further amplify the observed gains, but could also be more challenging to implement efficiently \citep{tan2024scatteredmixtureofexpertsimplementation}.

 We use the notation 1xFLOPs/2xFLOPs referring to the FLOPs \textit{in the MoE layer}: the latter variants activate twice the number of experts per token, hence proportionally increasing the computational cost. The actual difference in the total training FLOPs is smaller between the variants due to the use of non-MoE Transformer layers (e.g. Attention) and can be calculated based on the number of active parameters for each model (please refer to Table \ref{tab:10b_configs} \& Table \ref{tab:50b_configs} for exact numbers in the models we train).

 Please also note that many works use the Top-2 MoE routing (including, in the context of Transformers, GLaM \citep{du2022glam} or OpenMoE \citep{xue2024openmoeearlyeffortopen}). Here, we refer to this variant as Mixtral-like, since it was popularized by the release of the open-source Mixtral model \citep{jiang2024mixtral}, using Top-2 routing and 8 experts, and achieving the state-of-the-art performance at the time of release.

\paragraph{Training data.} Models are pretrained on random subsets with up to 300B tokens sampled from a large, diverse multilingual corpus containing both text and code (see \citet{parmar2024nemotron415btechnicalreport}). Before benchmark evaluation, we perform additional continued pretraining on a high-quality, filtered dataset containing alignment-style question-answer pairs. This step helps to improve benchmark signal by adapting models toward the evaluation distribution.

\paragraph{Training hyperparameters.} We use the AdamW \citep{loshchilov2019decoupledweightdecayregularization} optimizer with $\beta_1 = 0.9$, $\beta_2 = 0.95$, and weight decay set to 0.1. Models are trained with batches of 4M tokens and a sequence length of 2048. The learning rate is set to $2 \times 10^{-4}$ and follows a cosine schedule with linear warmup over the first $1\%$ of training steps. We initialize model weights with a standard deviation of $\sigma = 0.01$ and apply attention dropout with $p = 0.1$. The MoE capacity factor is set to 1.5, with auxiliary loss coefficient $1 \times 10^{-2}$ and z-loss coefficient $1 \times 10^{-3}$. Gradient clipping is used with a threshold of 1.0.
For the continued pretraining phase on the high-quality, filtered dataset, we generally follow the procedure in \citep{parmar2024reusedontretrainrecipe}. We resume from the model checkpoints without loading the optimizer states. The learning rate follows a cosine schedule, starting from $\eta_{\text{end}}$, the final learning rate of the initial phase, and decaying to $0.1 \eta_{\text{end}}$. Continued pretraining uses a token budget equal to $10\%$ of the original pretraining.

\subsubsection{Evaluation metrics} \label{sub:eval_metrics}

We evaluate the MoE variants from two complementary perspectives:

\textit{Efficiency gains.} To measure training efficiency, we track pretraining loss, a standard proxy for model quality \citep{kaplan2020scaling, hoffmann2022training, clark2022unified}. We report the validation loss of the models in plots and tables.

\textit{Downstream accuracy.} Beyond pretraining efficiency, we assess task-specific performance using a suite of established benchmarks: ARC-Easy, ARC-Challenge \citep{clark2018thinksolvedquestionanswering}, CommonsenseQA \citep{talmor2019commonsenseqaquestionansweringchallenge}, HellaSwag \citep{zellers2019hellaswagmachinereallyfinish}, MMLU \citep{hendrycks2021measuringmassivemultitasklanguage}, OpenBookQA \citep{mihaylov2018suitarmorconductelectricity}, PIQA \citep{bisk2019piqareasoningphysicalcommonsense}, RACE \citep{lai2017racelargescalereadingcomprehension}, SocialIQA \citep{sap2019socialiqacommonsensereasoningsocial}, TruthfulQA \citep{lin2022truthfulqameasuringmodelsmimic}, WinoGrande \citep{sakaguchi2019winograndeadversarialwinogradschema}. We use 5-shot evaluation for MMLU and zero-shot for all the other metrics.

\subsection{Comparing 11B Models} \label{sec:11b}
We begin our comparison by examining the effects of granularity and MoE design on models with approximately 11B total parameters, pretrained on 50B tokens. Model configurations are listed in Table \ref{tab:10b_configs}.

\vspace{0.3cm}

\begin{figure}
    \centering
    \subfigure[]{
        \includegraphics[width=0.3\textwidth]{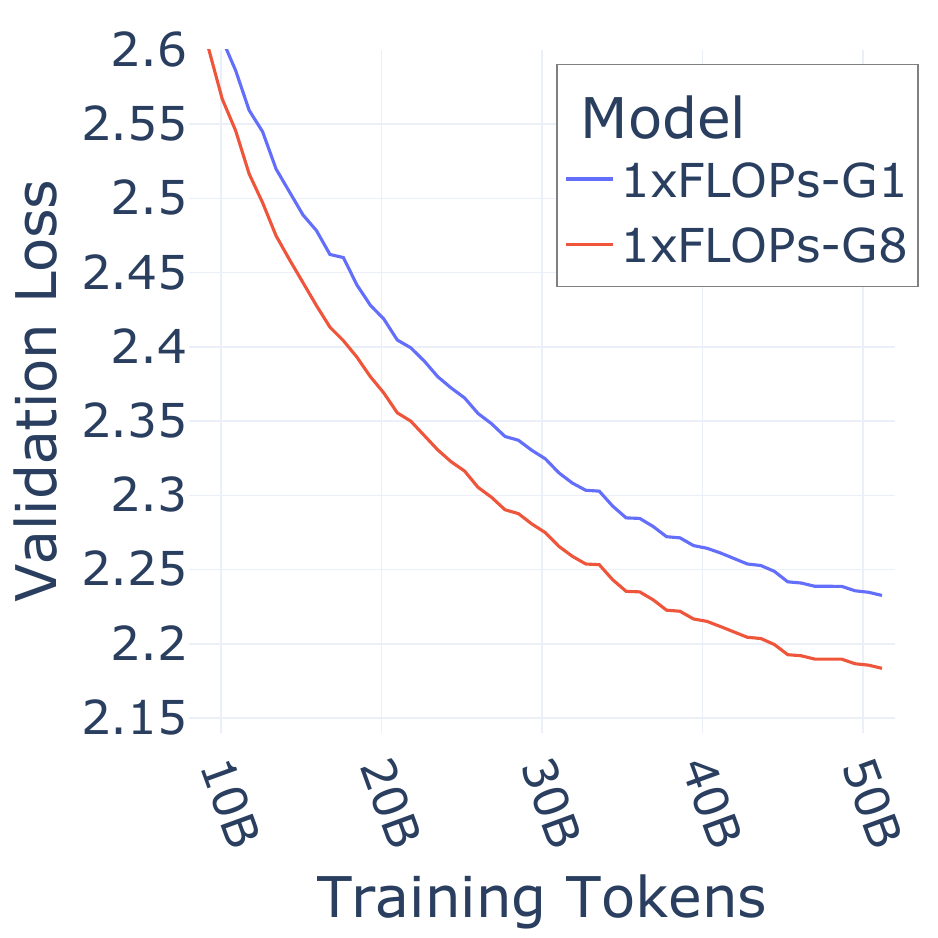}
    }
    \subfigure[]{
        \includegraphics[width=0.3\textwidth]{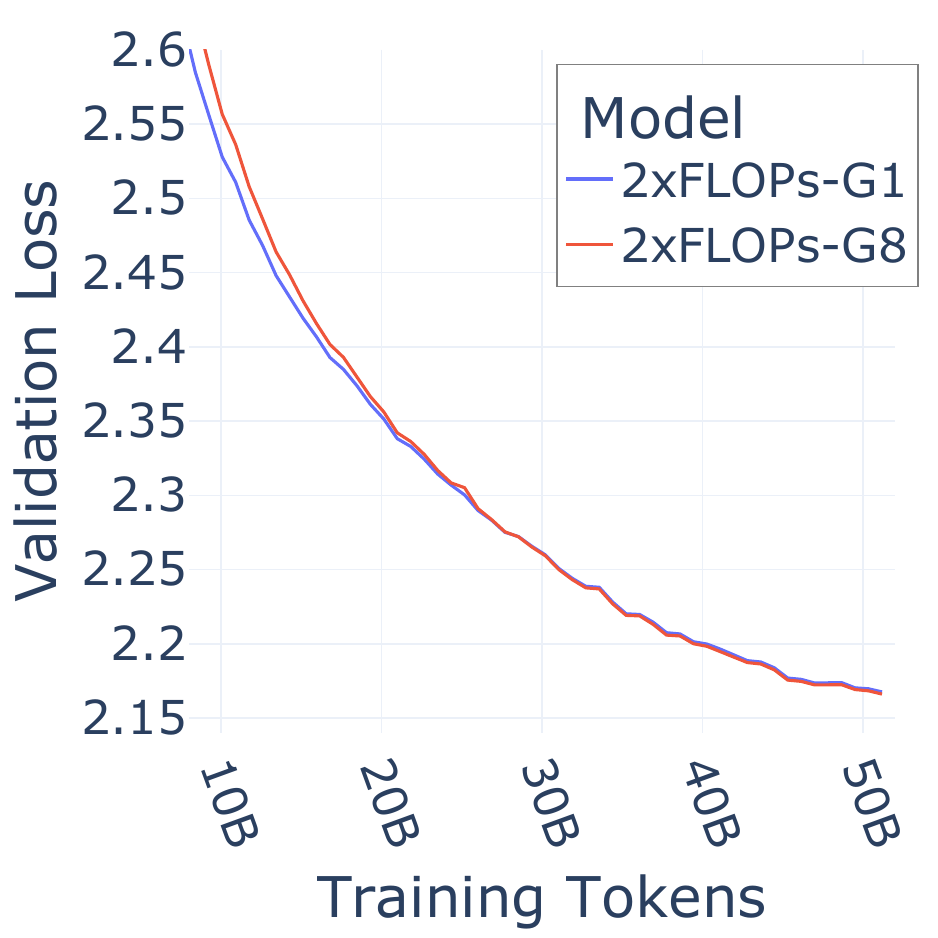}
    }
    \subfigure[]{
        \includegraphics[width=0.3\textwidth]{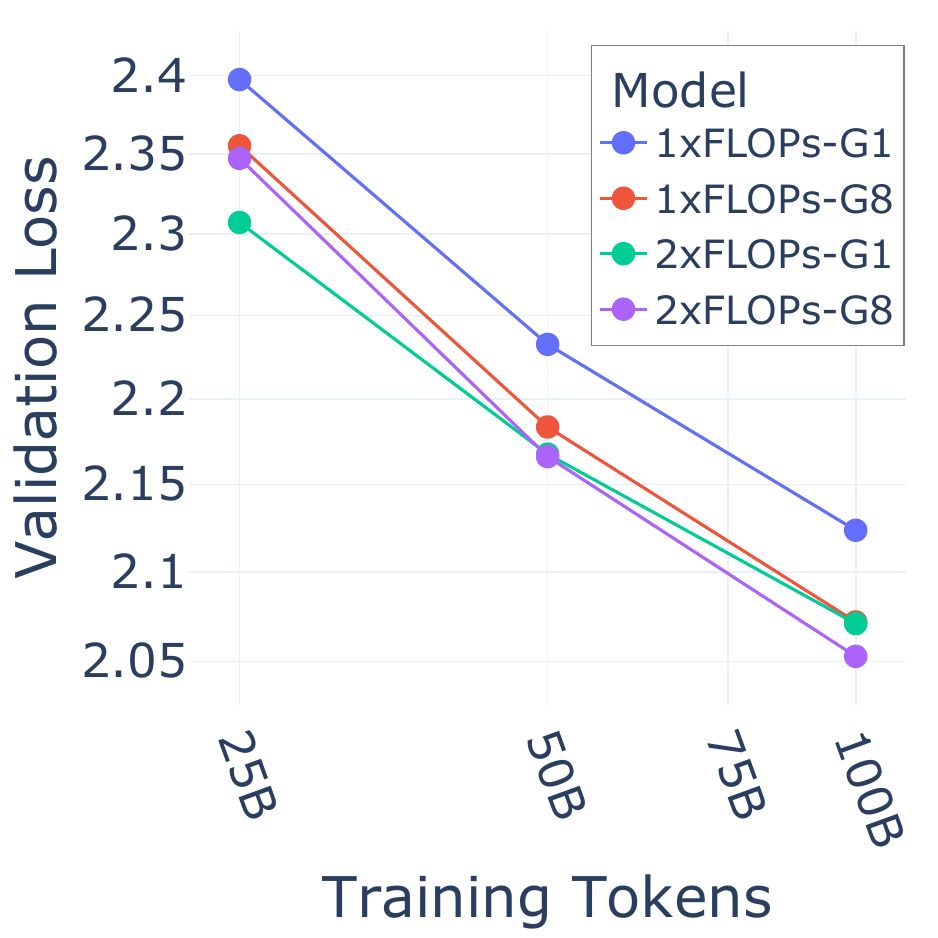}
    }
\caption{\textbf{(a)} - \textbf{(b)}: Validation loss curves of 11B models. Granularity improves performance of the Switch model, but doesn't bring advantage for the Mixtral variant. This observation changes with longer training. \textbf{(c)}: Comparison of the final loss for scaling the training length of the 11B models. Fine-grained variants perform relatively the best on the longest token horizon.}
    
  \label{fig:train_length}
\end{figure}

\begin{table}[ht!]
\caption{Hyperparameter configurations for 11B models.} 
\label{tab:10b_configs}
\centering
\resizebox{\textwidth}{!}{
\begin{tabular}{ccccccc}
\toprule
\multirow{2}{*}{Name} & \multirow{2}{*}{Experts} & \multirow{2}{*}{Router Top-k} & \multirow{2}{*}{Active Parameters} & \multirow{2}{*}{$d_\text{model}$} & \multirow{2}{*}{$d_\text{expert}$} & \multirow{2}{*}{Total Parameters} \\ \\
\midrule
1xFLOPs-G1 & 8 & 1 & 2.7B & 2048 & 8192 & 11.1B \\
\midrule
1xFLOPs-G8 & 64 & 8 & 2.7B & 2048 & 1024 & 11.1B \\
\midrule
2xFLOPs-G1 & 8 & 2 & 3.9B & 2048 & 8192 & 11.1B \\
\midrule
2xFLOPs-G8  & 64 & 16 & 3.9B & 2048 & 1024 & 11.1B \\
\bottomrule
\end{tabular}
}
\end{table}

\vspace{0.3cm}

Fig. \ref{fig:train_length} (a)-(b) depicts the loss curves of the models, while Table~\ref{tab:10b_res} presents the validation loss and downstream benchmark results for these four configurations. Below we analyze the impact of granularity within each FLOPs category (1x and 2x).

\vspace{0.2cm}

Comparing the Switch-style variants, we observe that adding granularity (1xFLOPs-G8) significantly improves accuracy over the standard configuration (1xFLOPs-G1). This result proves the advantage of using fine-grained experts in this setup. 

\vspace{0.2cm}

As expected, the standard Mixtral-style model (2xFLOPS-G1), activating two experts per token, outperforms the standard Switch-style model (1xFLOPS-G1), which activates only one. This reflects the benefit of increasing the number of parameters activated per each token. 

\vspace{0.2cm}

Interestingly, when comparing the 2xFLOPs variants, adding granularity (2xFLOPS-G8) does not yield a similarly significant improvement over the standard configuration (2xFLOPS-G1). As seen in Fig.~\ref{fig:train_length}(b), the validation losses of both models are very close, although we still see a modest improvement in most benchmark scores for the fine-grained architecture. In Sec.~\ref{sec:longer} \& Sec.~\ref{sec:50b} we show that the advantage of fine-grained Mixtral-like variant is more pronounced with sufficiently long training and larger models.

\pagebreak

\begin{table}[h!]
\caption{Evaluation results for the 11B models. Within each pair (Switch-style, Mixtral-style), we mark the model with a better score for each metric.}
\label{tab:10b_res}
\centering
\resizebox{0.8\textwidth}{!}{
\begin{tabular}{lcccc}
\toprule
Benchmark & 1$\times$FLOPs-G1 & 1$\times$FLOPs-G8 & 2$\times$FLOPs-G1 & 2$\times$FLOPs-G8 \\
\midrule
ARC-C               & 29.7 & \textbf{32.7} & 31.4 & \textbf{32.8} \\
ARC-E               & 58.1 & \textbf{60.1} & 60.6 & \textbf{62.3} \\
HellaSwag           & 53.0 & \textbf{57.4} & 56.8 & \textbf{58.3} \\
OpenbookQA          & 32.0 & \textbf{33.8} & 31.6 & \textbf{33.2} \\
PIQA                & 71.9 & \textbf{73.0} & \textbf{75.0} & 74.2 \\
SocialIQA           & 41.6 & \textbf{42.6} & 42.6 & \textbf{44.3} \\
WinoGrande          & 52.6 & \textbf{54.4} & \textbf{55.7} & 55.5 \\
\midrule
Average             & 48.4 & \textbf{50.6} & 50.5 & \textbf{51.5} \\
\midrule
Valid Loss          & 2.233 & \textbf{2.183} & 2.168 & \textbf{2.166} \\
\bottomrule
\end{tabular}}
\end{table}

These initial 11B model results suggest that at this scale and training duration:

\begin{itemize}
    \item In the Switch-like setup it is beneficial to replace standard experts with fine-grained ones. 
    \item Mixtral-like model achieves better quality on a given dataset than the Switch one. This is achieved at the cost of activating more parameters per each token.
    \item Adding granularity on top of the Mixtral model does not bring significant advantage. In the later sections, we show this observation changes with longer training and larger models.
\end{itemize}

In the following sections we analyze how changing the dataset and model size influences these results.

\subsection{The Effect of Training Length} \label{sec:longer}

\begin{table}[h!]
\caption{Training step savings measured in \% of training steps needed to reach the final loss of the standard Switch-MoE (1xFLOPs-G1). The savings increase for fine-grained MoE and remain constant for the standard Mixtral architecture.} 
\label{tab:savings_length}
\centering
\resizebox{0.6\textwidth}{!}{
\begin{tabular}{cccc}
\toprule
\multirow{2}{*}{Name} & \multirow{2}{*}{25B tokens} & \multirow{2}{*}{50B tokens} & \multirow{2}{*}{100B tokens} \\ \\
\midrule
1xFLOPs-G8 & 21.6\% & 27.9\% & 33.6\% \\
\midrule
2xFLOPs-G1 & 34.5\% & 32.8\% & 32.8\% \\
\midrule
2xFLOPs-G8  & 24.6\% & 32.8\% & 39.4\% \\
\bottomrule
\end{tabular}
}
\end{table}

Modern LLMs are pretrained on very large datasets and understanding how the training length affects different MoE configurations can provide an important context to our analysis.
To examine scaling the dataset size, we perform additional experiments training the 11B models on two token horizons - 25B and 100B tokens - alongside the original setup. Fig.~\ref{fig:train_length}(c) depicts the validation losses for these three training lengths. Notably, we observe more pronounced performance gains of fine-grained MoE as the dataset size grows.

At the shorter horizon of 25B tokens, the two fine-grained variants show validation losses similar to each other and fall between the two standard baselines. Surprisingly, the 2xFLOPs-G8 variant offers only a very modest improvement over 1xFLOPs-G8, even though it activates twice as many experts per token.

This picture changes when comparing models on the longest token horizon. In that setup, 1xFLOPs-G8 matches 2xFLOPs-G1. Notably, this means that the 1xFLOPs-G8 model achieves similar performance to 2xFLOPs-G1, despite activating just about half the MoE parameters. This results in cheaper training, but also a more lightweight inference. The 2xFLOPs-G8 exhibits the best validation loss among the considered setups.

Directly comparing absolute pretraining loss values can be difficult. To provide a more intuitive measure of efficiency gains, in Table~\ref{tab:savings_length} we report \textit{training step savings}. This metric quantifies how many fewer training steps a specific variant needs to reach the same final validation loss as the baseline Switch model. We report savings as a percentage of the total training steps. Examining the results, we can see that the standard Mixtral model shows a roughly constant advantage among the three dataset sizes. In contrast, the savings from using fine-grained MoE are growing with the training duration. This suggests that the granular models become increasingly efficient with more training data.

These results highlight the importance of considering the training length in the analysis, with fine-grained models especially benefiting from extended token budgets.

\subsection{Scaling Beyond 50B Model Parameters} \label{sec:50b}
We further scale our analysis by comparing a similar set of models this time with 56B total parameters, trained on 300B tokens. The key architecture hyperparameters are presented in Table \ref{tab:50b_configs}.

\begin{table}[ht!]
\caption{Hyperparameter configurations for the largest models.} 
\label{tab:50b_configs}
\centering
\resizebox{\textwidth}{!}{
\begin{tabular}{ccccccc}
\toprule
\multirow{2}{*}{Name} & \multirow{2}{*}{Experts} & \multirow{2}{*}{Router Top-k} & \multirow{2}{*}{Active Parameters} & \multirow{2}{*}{$d_\text{model}$} & \multirow{2}{*}{$d_\text{expert}$} & \multirow{2}{*}{Total Parameters} \\ \\
\midrule
1xFLOPs-G1 & 8 & 1 & 10.7B & 4096 & 16384 & 55.8B \\
\midrule
1xFLOPs-G8 & 64 & 8 & 10.7B & 4096 & 2048 & 55.8B \\
\midrule
2xFLOPs-G1 & 8 & 2 & 17.1B & 4096 & 16384 & 55.8B \\
\midrule
2xFLOPs-G8  & 64 & 16 & 17.1B & 4096 & 2048 & 55.8B \\
\bottomrule
\end{tabular}
}
\end{table}

\begin{figure}
    \centering
    \subfigure{
        \includegraphics[width=0.48\textwidth]{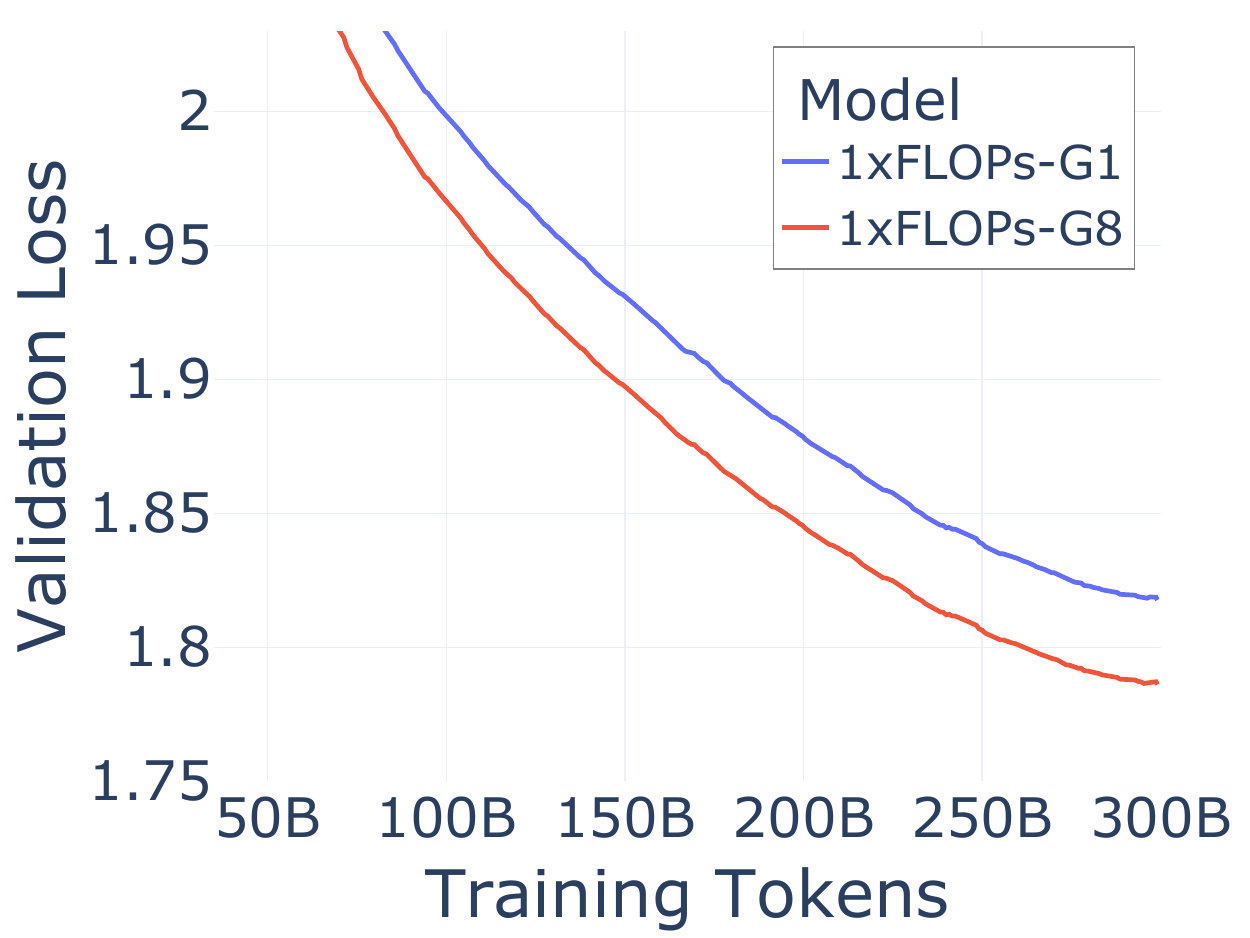}
    }
    \subfigure{
        \includegraphics[width=0.48\textwidth]{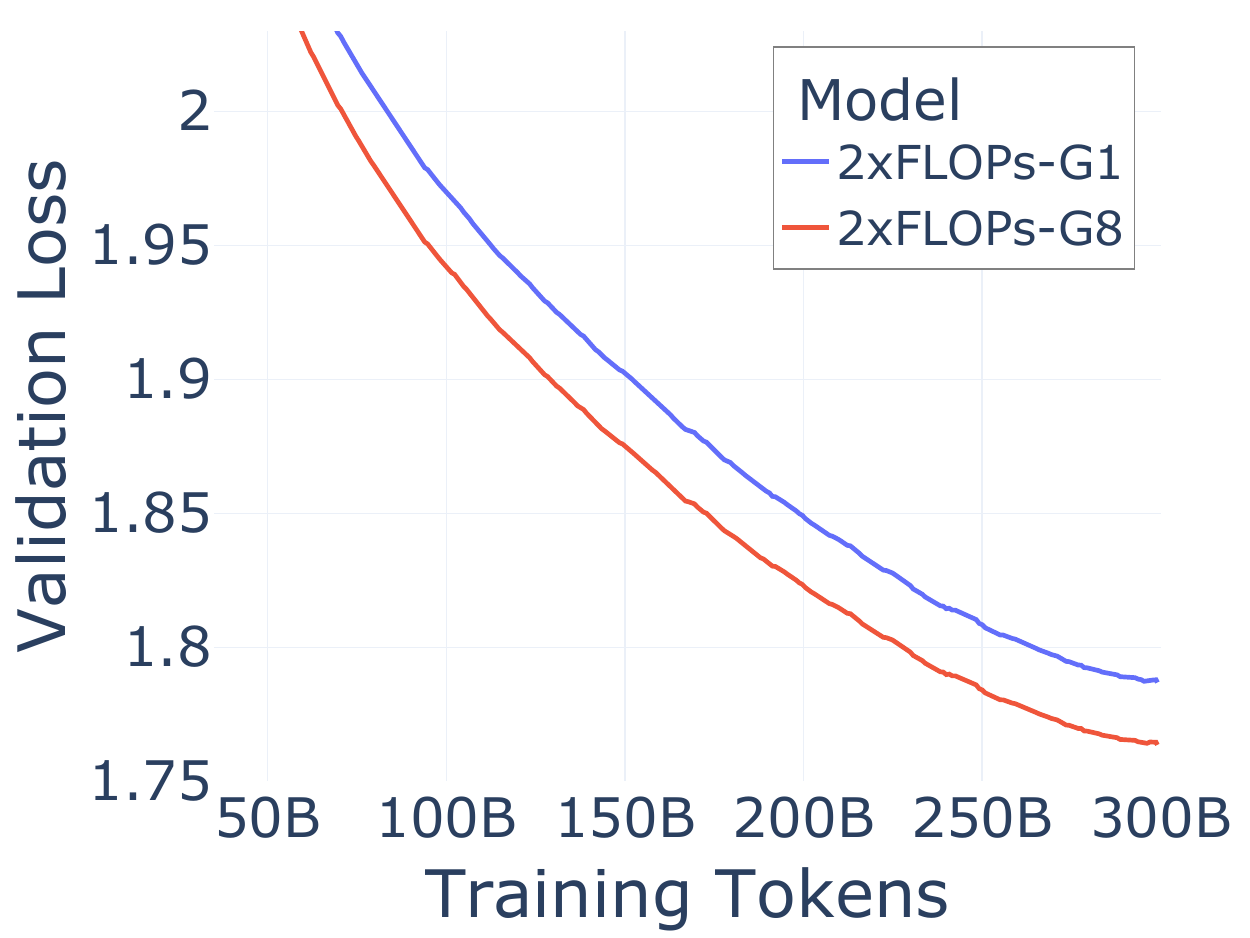}
    }
\caption{
Validation loss curves of 56B models. \textit{(left)}: Standard MoE (Switch variant, 1xFLOPs-G1) and its fine-grained counterpart. \textit{(right):} Standard MoE (Mixtral variant, 2xFLOPs-G1)
and its fine-grained counterpart. In both cases we show gains from increasing expert granularity.
}
    
  \label{fig:loss_56b}
\end{figure}

Loss curves of the models are presented in Fig. \ref{fig:loss_56b}. We observe the improvement from using fine-grained MoE in both Switch and Mixtral variants. Consequently, the relative performance ranking of the considered architectures closely resembles the order observed for the 11B models on the longest token horizon (Sec. \ref{sec:longer}). The 1xFLOPs-G8 variant outperforms the standard Switch MoE and matches the more computationally expensive 2xFLOPs-G1. Adding granularity on top of the Mixtral variant results in the best performance across all models.

Table \ref{tab:eval_largest} shows how these observations translate to downstream evaluations. We see the same relationships in performance as observed for validation loss, with only minor variations on specific metrics.

\definecolor{PlotlyBlueRGB}{rgb}{0.388,0.431,0.980}

\begin{table}[h!]
\caption{Evaluation results for the 56B models. Within each pair (Switch-style, Mixtral-style), we mark the model with a better score for each metric.}
\label{tab:eval_largest}
\centering
\resizebox{0.8\textwidth}{!}{
\begin{tabular}{lcccc}
\toprule
Benchmark & 1$\times$FLOPs-G1 & 1$\times$FLOPs-G8 & 2$\times$FLOPs-G1 & 2$\times$FLOPs-G8 \\
\midrule
ARC-C               & 44.3 & \textbf{46.4} & 45.6 & \textbf{47.7} \\
ARC-E               & 73.2 & \textbf{76.1} & 74.3 & \textbf{77.6} \\
CommonsenseQA       & 60.4 & \textbf{65.3} & 67.6 & \textbf{68.7} \\
HellaSwag           & 75.0 & \textbf{77.0} & 76.1 & \textbf{78.1} \\
MMLU                & 47.5 & \textbf{50.1} & 50.5 & \textbf{52.1} \\
OpenbookQA          & \textbf{42.6} & 42.0 & 42.0 & \textbf{45.0} \\
PIQA                & 80.5 & \textbf{81.0} & 79.9 & \textbf{81.2} \\
RACE                & 58.1 & \textbf{60.7} & \textbf{60.8} & 59.3 \\
SocialIQA           & 47.4 & \textbf{47.7} & 46.3 & \textbf{46.9} \\
TruthfulQA          & 34.2 & \textbf{35.2} & 36.3 & \textbf{38.8} \\
WinoGrande          & 66.9 & \textbf{67.5} & 67.9 & \textbf{70.2} \\
\midrule
Average             & 57.3 & \textbf{59.0} & 58.8 & \textbf{60.5} \\
\midrule
Valid Loss          & 1.811 & \textbf{1.779} & 1.780 & \textbf{1.757} \\
\bottomrule
\end{tabular}}
\end{table}

\pagebreak

To summarize, on our largest 56B models:
\begin{itemize}
    \item We continue to see the performance improvements when using fine-grained MoE in the Switch (1xFLOPs) setup.
    \item 1xFLOPs-G8 not only outperforms 1xFLOPs-G1, but also matches 2xFLOPs-G1. This is true despite the fine-grained variant activating fewer parameters, hence being more compute efficient for training and inference.
    \item As expected, standard MoE with Top-2 choice (2xFLOPs-G1) achieves better results after training on a given dataset than the Top-1 version (1xFLOPs-G1).
    \item The Mixtral variant with granularity (2xFLOPs-G8) performs better than its standard counterpart (2xFLOPs-G1). 
\end{itemize}

\section{Discussion on MoE Design and Training} \label{sec:analysis_moe}

In this section, we investigate several important design choices and training behaviors that affect MoE performance. We begin by examining expert load imbalance (Sec.~\ref{sub:expert_load}). Next, we explore how the router’s logit magnitude evolves throughout training of fine-grained MoE (Sec. \ref{sub:logits_magnitude}), before concluding with an analysis on the optimal order of softmax and Top-k placement in the router (Sec. \ref{sub:soft_topk_order}). The models described in this section follow the training setup of Sec.~\ref{sec:11b} (11B total parameters, 50B training tokens).

\subsection{Expert Load Imbalance} \label{sub:expert_load}

\begin{figure}[h!]
    \centering
    \subfigure{
        \includegraphics[width=0.48\textwidth]{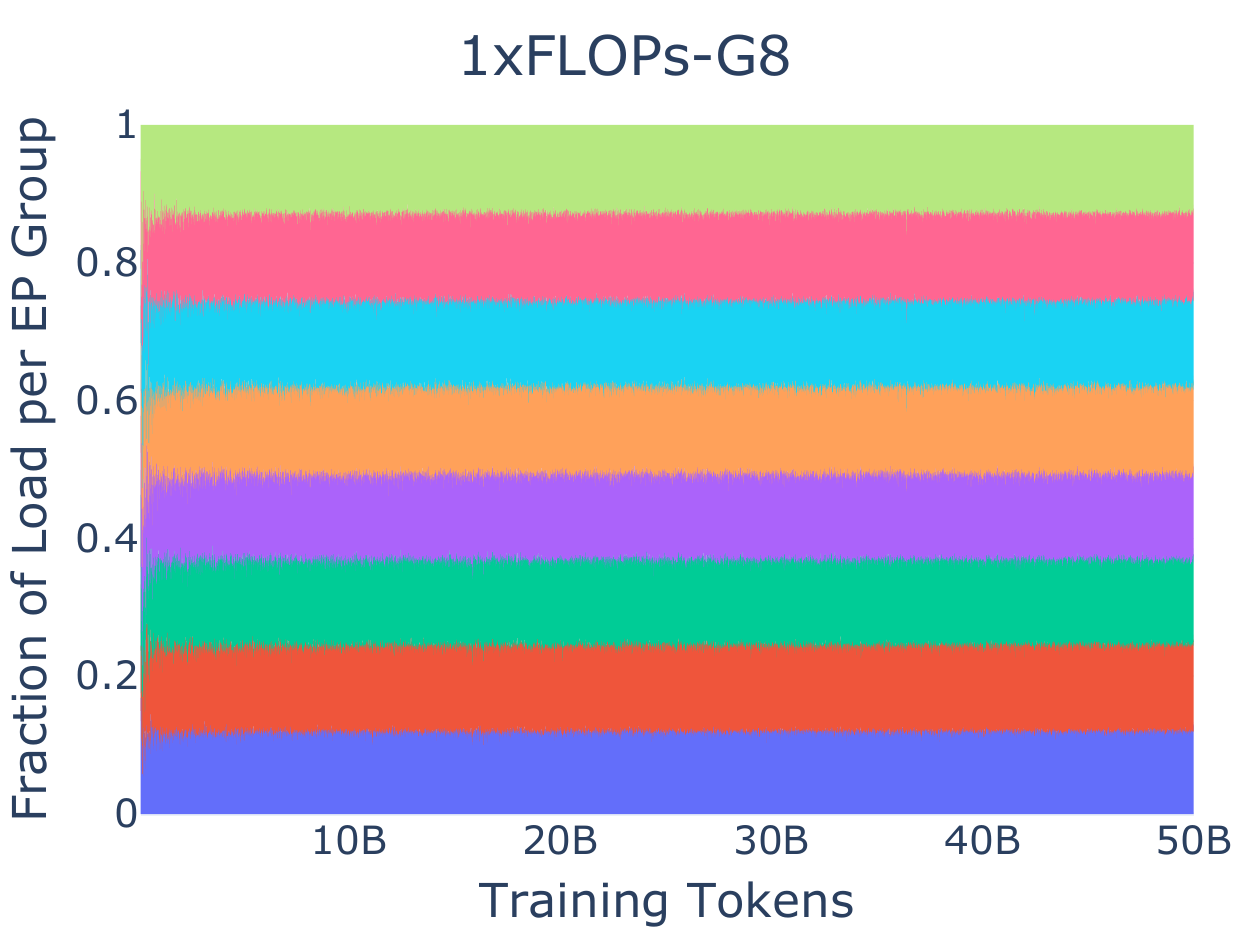}
    }
    \subfigure{
        \includegraphics[width=0.48\textwidth]{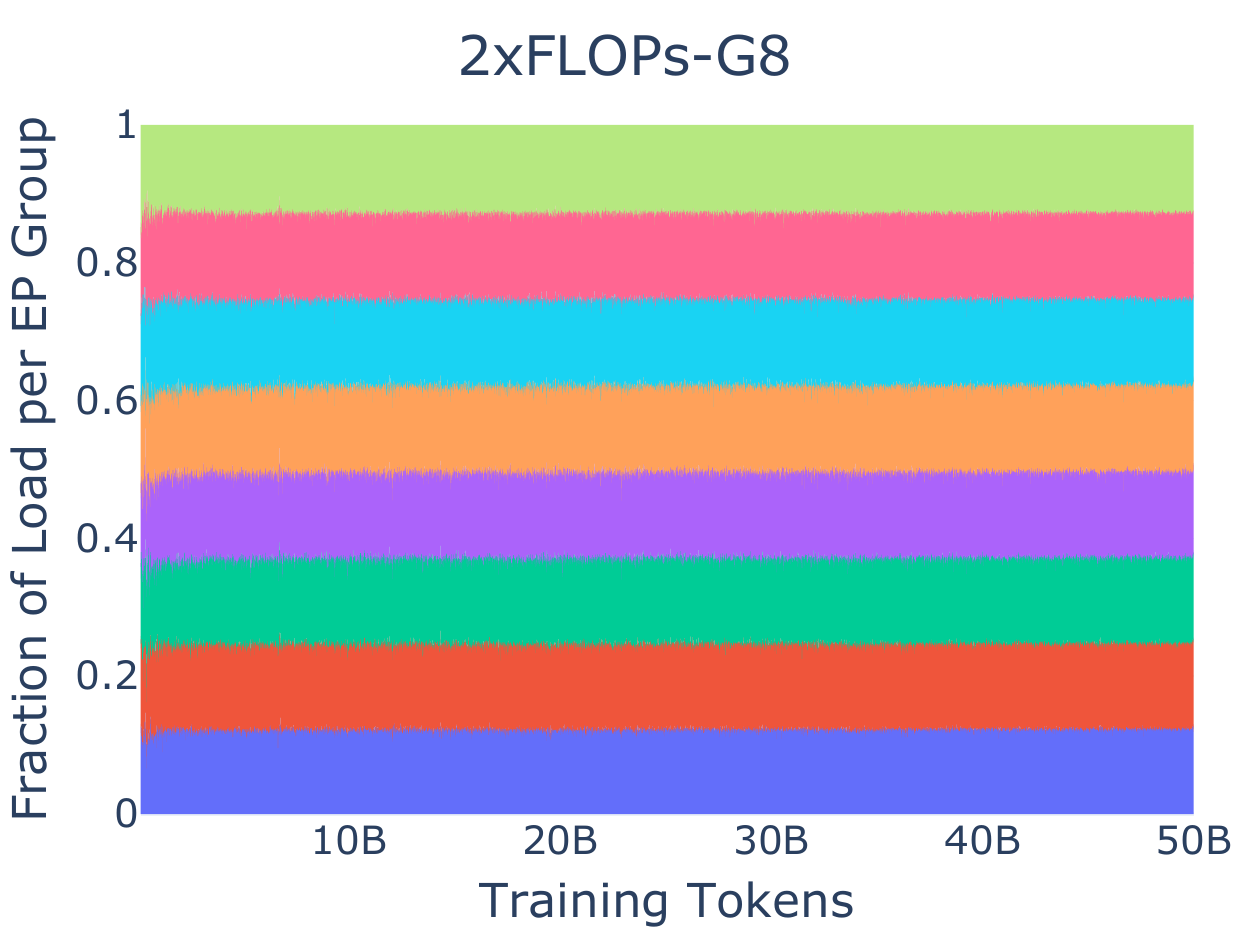}
    }
\caption{Fraction of load assigned to each Expert Parallel group for fine-grained models. We generally observe a balanced assignment of tokens between different EP groups. This is similar in the middle and final layer of the model (see App. \ref{app:imbalance}).}
    
  \label{fig:ep_load}
\end{figure}

In the standard (Token Choice) MoE architecture \citep{shazeer2017outrageously, fedus2022switch}, routing scores are calculated independently for each token. This independence means expert assignments are not guaranteed to be balanced. As a result, some experts might be overloaded with tokens while others remain underutilized. To prevent this imbalance, two techniques are commonly used \citep{fedus2022switch}: load balancing loss and capacity factor. Load balancing loss encourages a more even distribution of tokens across experts. Capacity factor limits the maximum number of tokens each expert can process. If an expert receives more tokens than its capacity allows, these excess tokens are dropped.

Fine-grained MoE architectures, where each token is routed to more experts chosen from a larger overall pool, could potentially increase this imbalance risk. This becomes particularly important when using expert parallelism, distributing experts across different devices (e.g., GPUs). A crucial goal in such distributed settings is to ensure each device processes roughly the same number of tokens per forward pass, avoiding delays caused by waiting for the most heavily loaded device. The mapping of experts to devices is determined by the Expert Parallel (EP) size. For our setup, using a fine-grained MoE with 64 experts and an EP size of 8 means each device is responsible for 8 experts.

To assess how well the load was balanced across devices in our fine-grained models, we tracked the fraction of the total token load processed by each of the 8 EP groups during training. These measurements are shown for the initial layer in Fig.~\ref{fig:ep_load}, and for the middle and last layers in Figs.~\ref{fig:ep_load_app}~\&~\ref{fig:ep_load_app_2} (App.~\ref{app:imbalance}). The data generally indicates that the load balancing loss quickly leads to uniform load distribution across the EP groups.

\subsection{Evolution of the Router Logits Magnitude} \label{sub:logits_magnitude}

\begin{figure}[h!]
    \centering
    \subfigure{
        \includegraphics[width=0.17\textwidth]{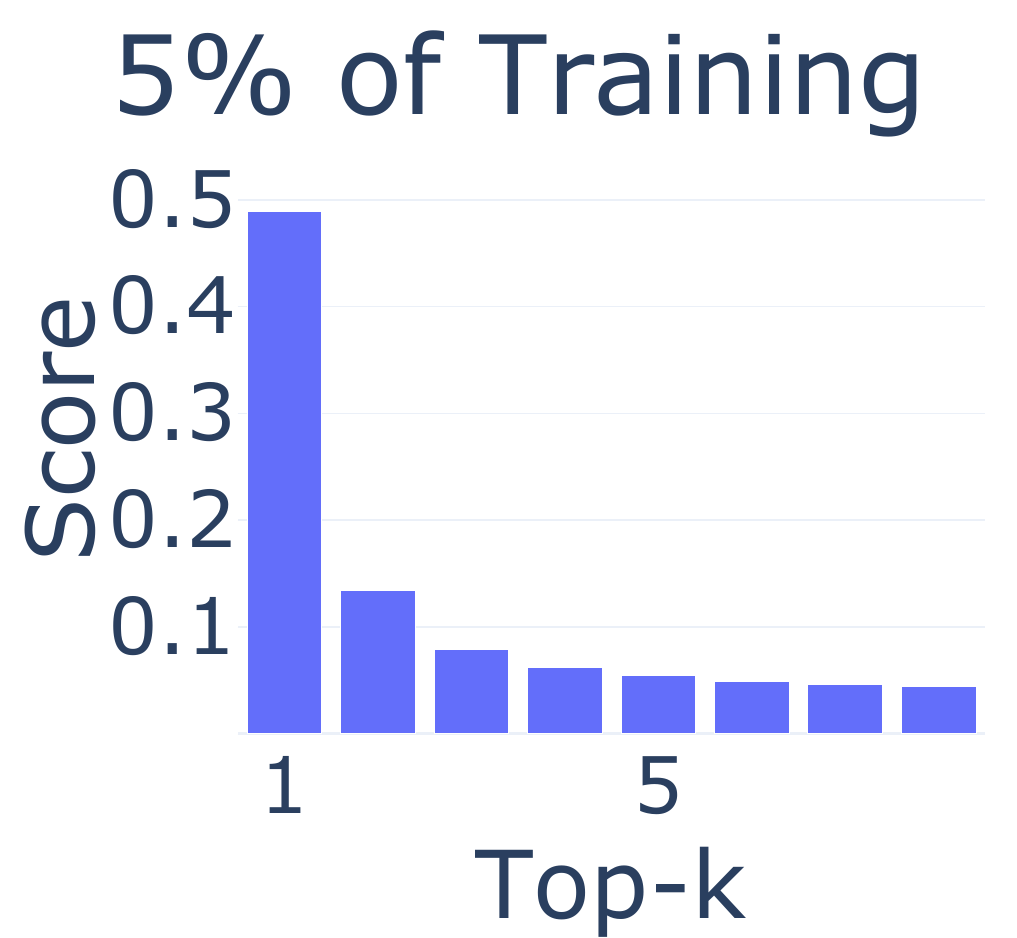}
    }
    \subfigure{
        \includegraphics[width=0.17\textwidth]{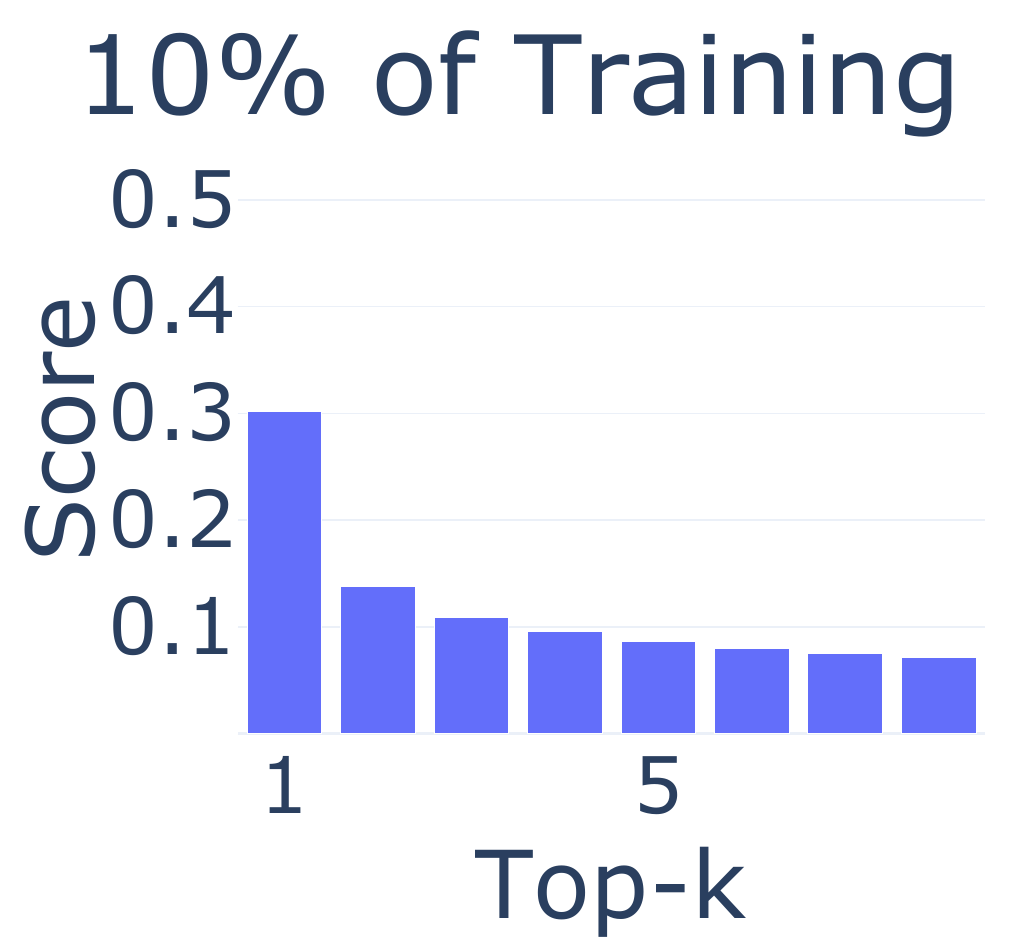}
    }
    \subfigure{
        \includegraphics[width=0.17\textwidth]{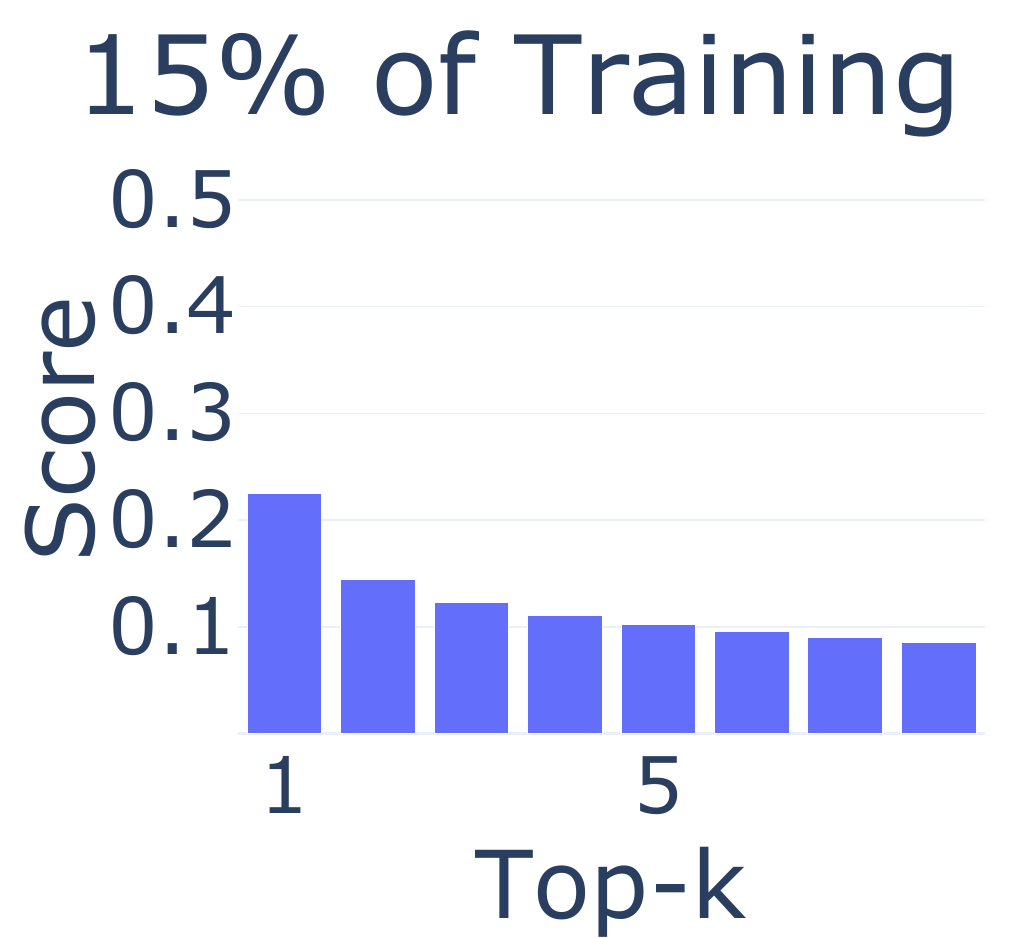}
    }
    \subfigure{
        \includegraphics[width=0.17\textwidth]{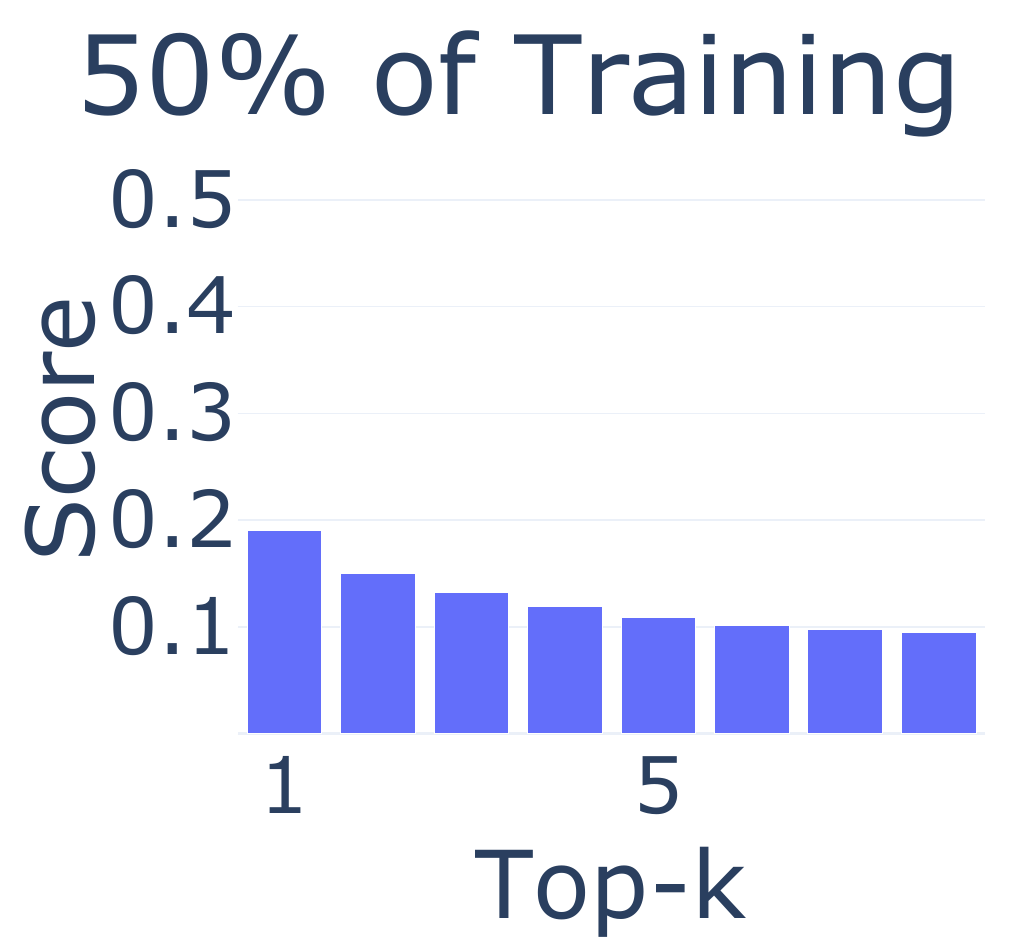}
    }
    \subfigure{
        \includegraphics[width=0.17\textwidth]{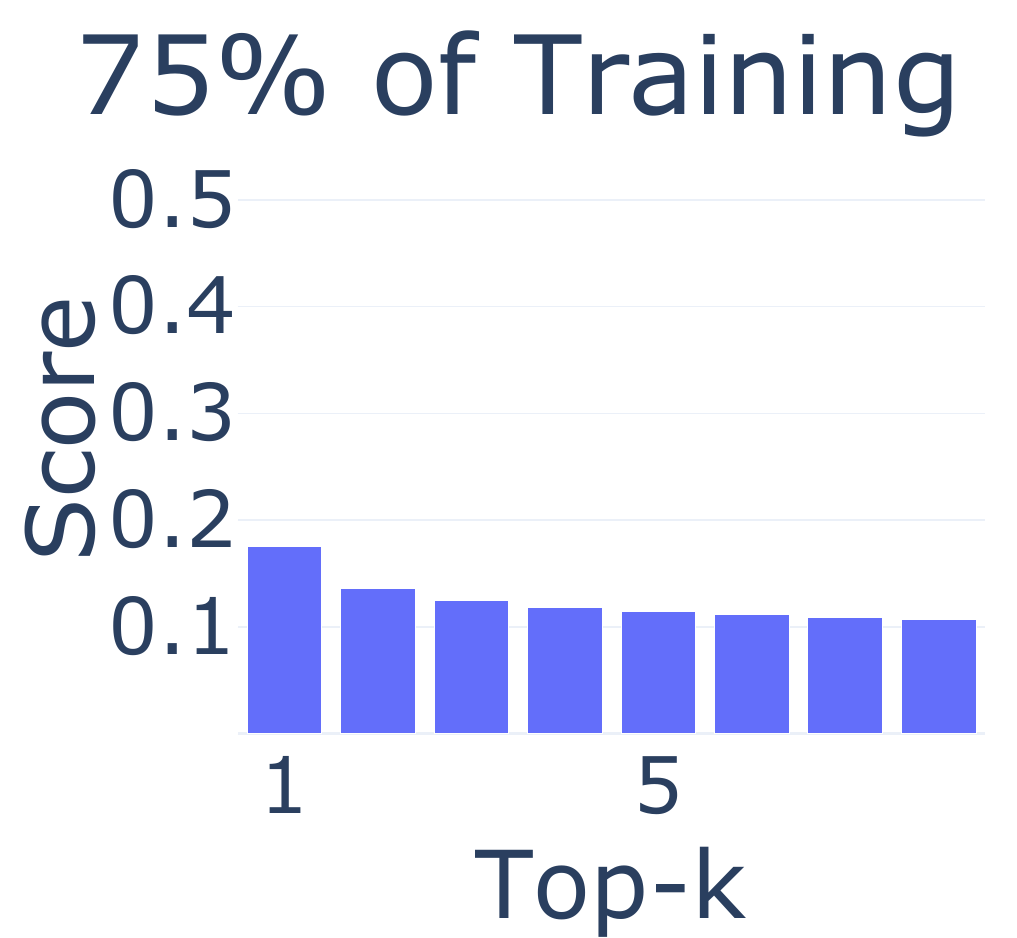}
    }\\
\small\textit{top}: 1xFLOPs-G8

\par\vspace{0.2em} 

    \subfigure{
        \includegraphics[width=0.17\textwidth]{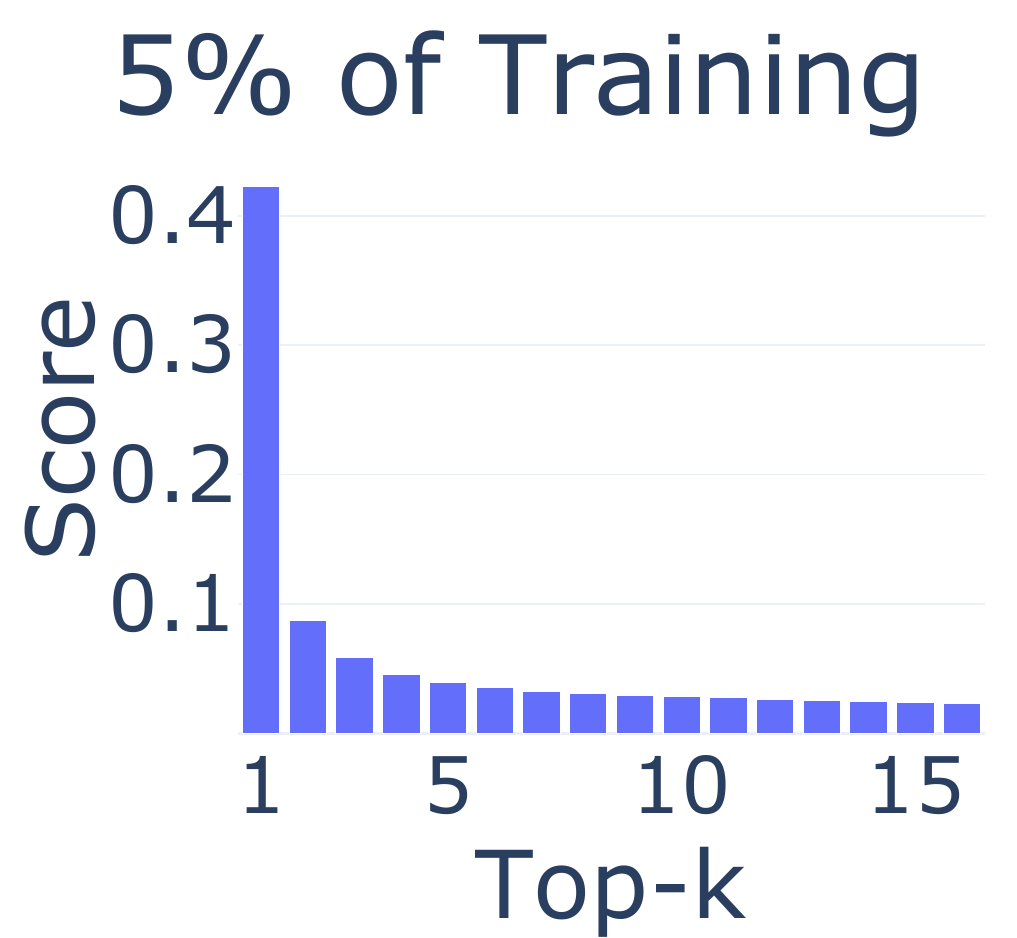}
    }
    \subfigure{
        \includegraphics[width=0.17\textwidth]{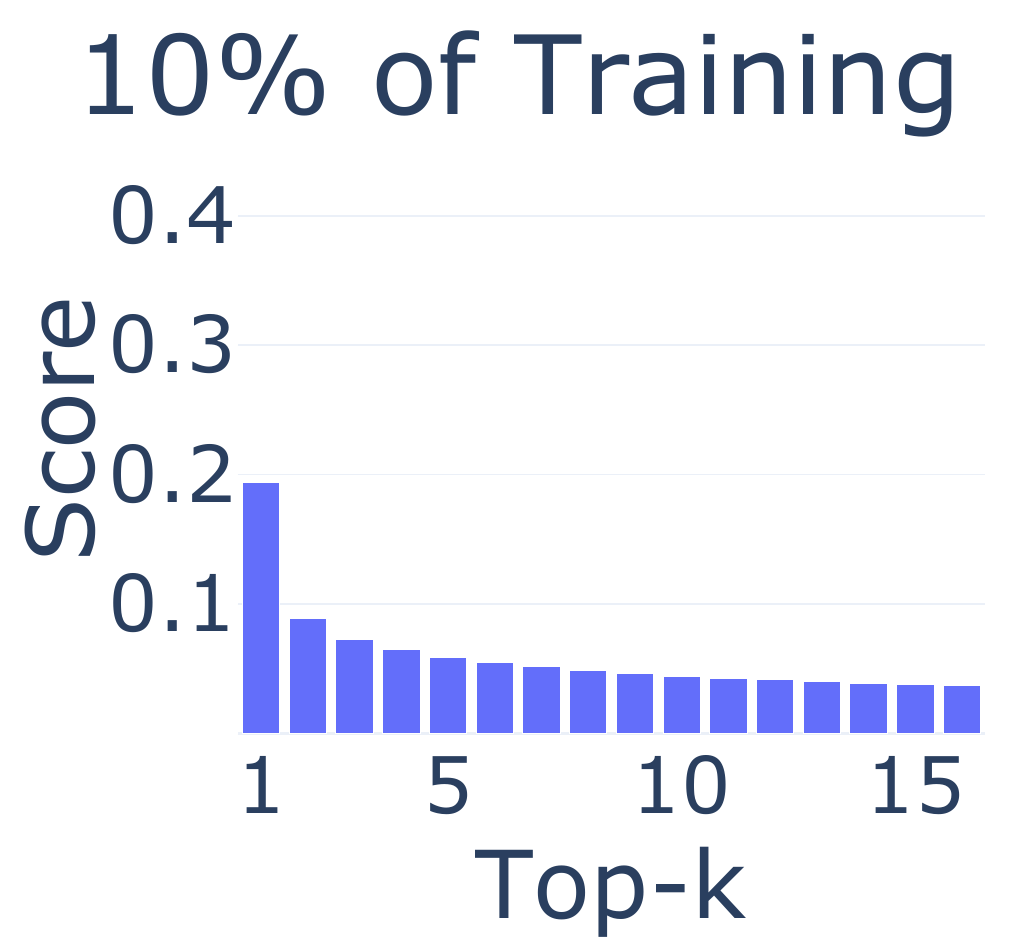}
    }
    \subfigure{
        \includegraphics[width=0.17\textwidth]{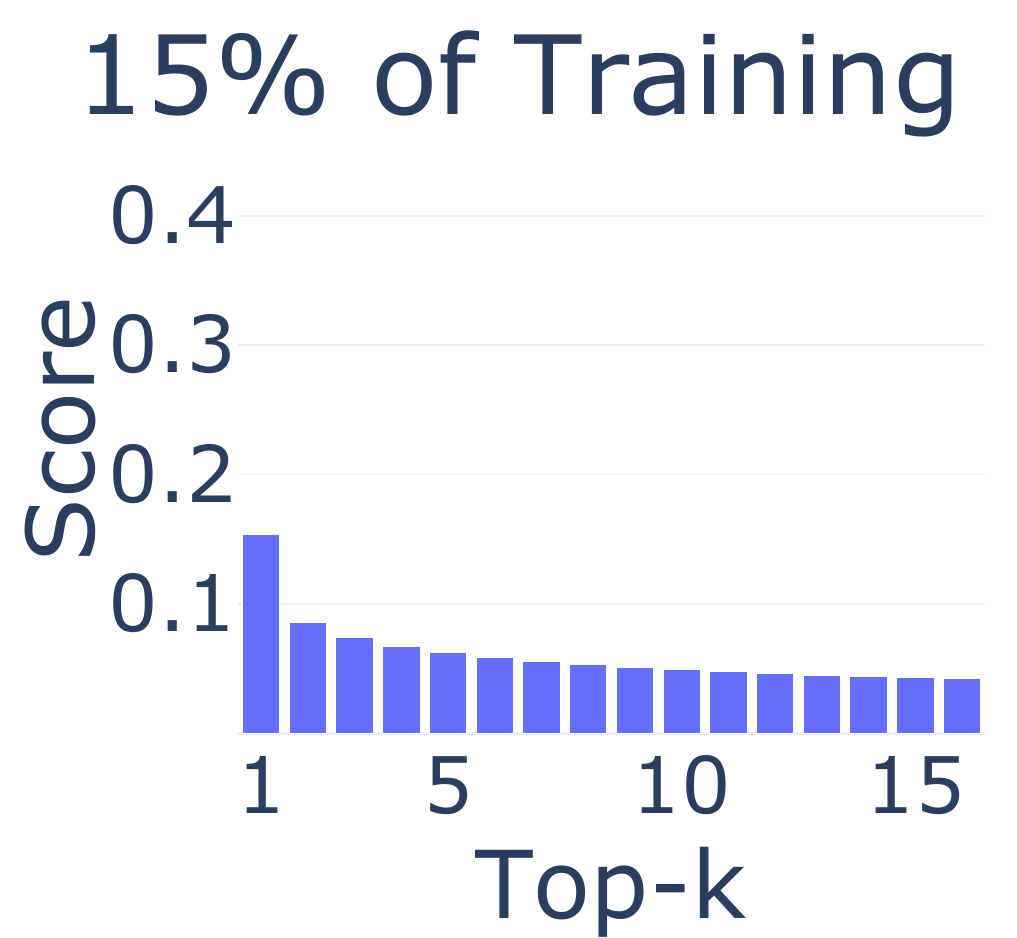}
    }
    \subfigure{
        \includegraphics[width=0.17\textwidth]{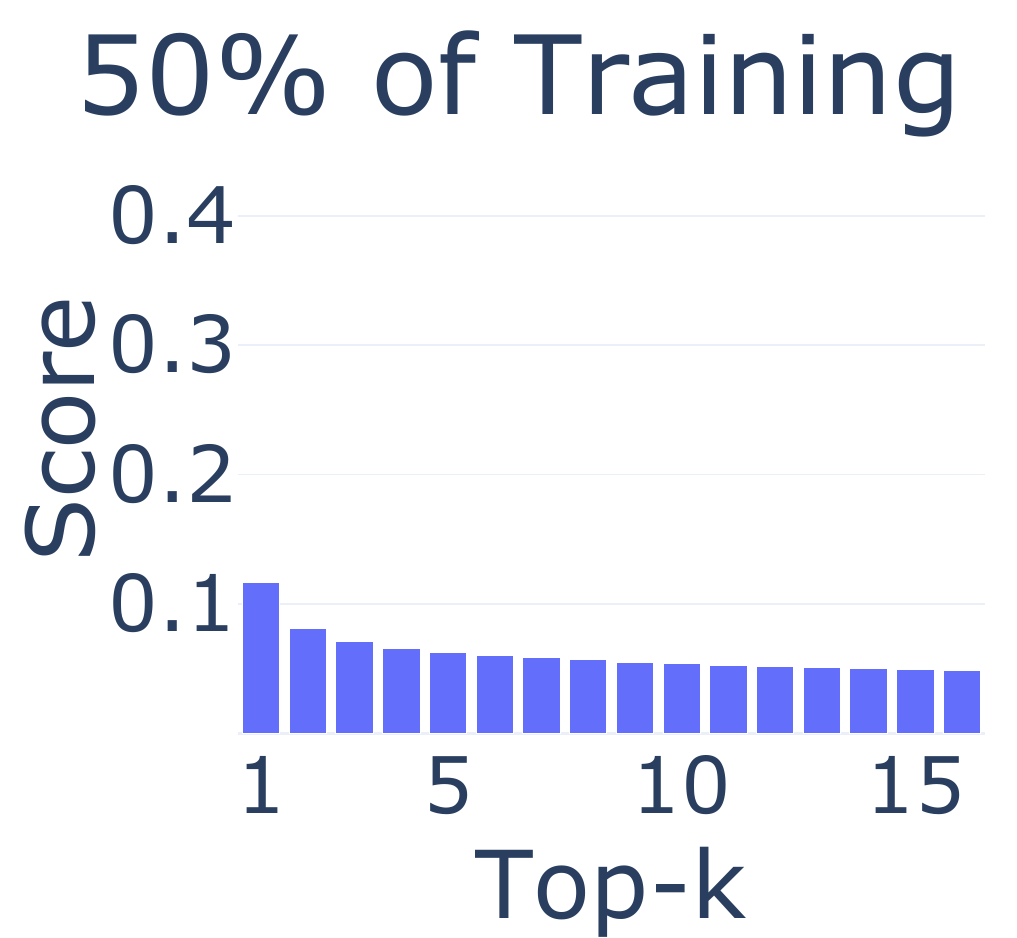}
    }
    \subfigure{
        \includegraphics[width=0.17\textwidth]{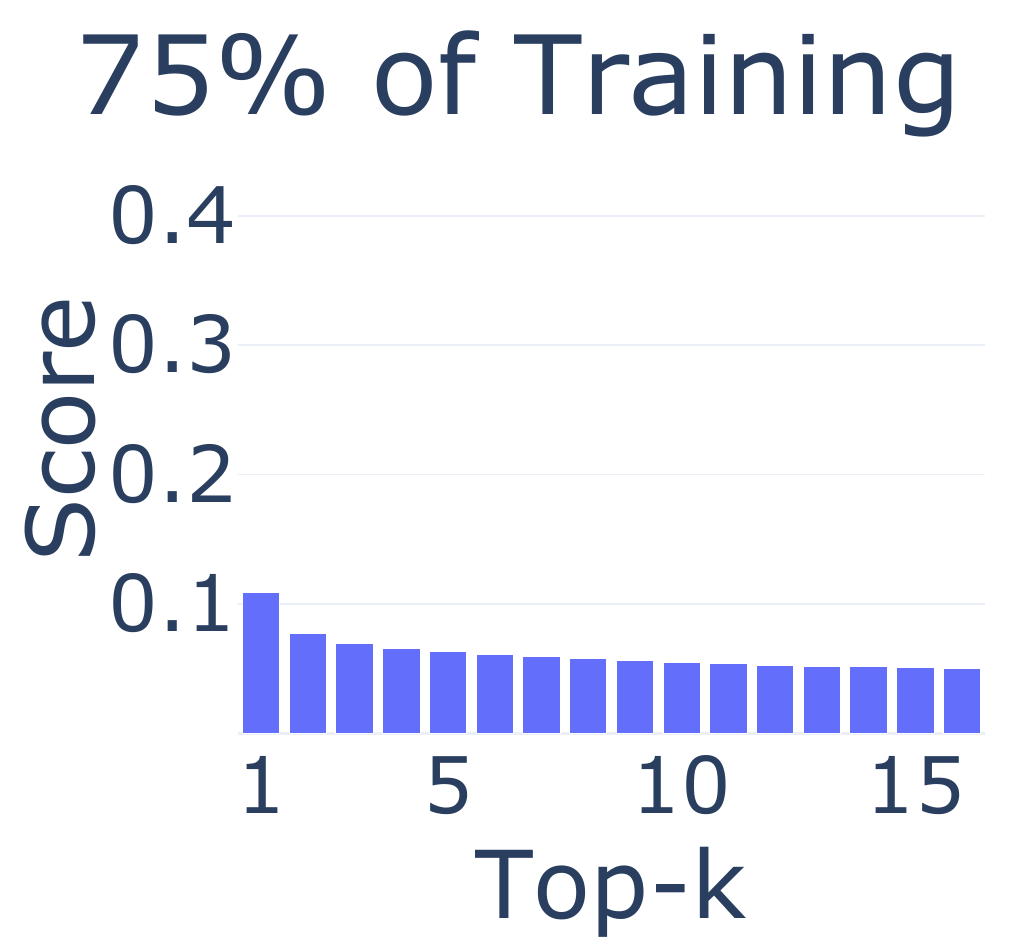}
    }\\
\small\textit{bottom}: 2xFLOPs-G8
\caption{ The distribution of the Top-k router logits in the initial layer of fine-grained models at selected points of training. In the beginning, the router assigns most of the weight to single expert, while gradually learning to use the remaining ones throughout training. We observe a similar pattern in the middle and last layers of the model (see App. \ref{app:logits}).}
    
  \label{fig:logit_distrib}
\end{figure}

As shown in Section~\ref{sec:longer}, the performance advantage of fine-grained MoE models increases with the amount of training data. While fine-grained MoE eventually outperforms its standard counterparts, its relative gain is less pronounced on shorter training horizons. Here, we explore a potential reason for this effect.

As shown in Fig.~\ref{fig:train_length}(c), on the shortest training horizon, the 2xFLOPs-G8 model shows little advantage over 1xFLOPs-G8, despite activating twice as many experts per token. This observation suggests a potential difficulty in router learning: the router may initially struggle to effectively utilize all the selected experts, focusing primarily on those with the highest scores. Consequently, simply increasing the number of activated experts (Top-k) provides limited benefit early in training if the router cannot leverage them.

\vspace{0.4cm}

To test this hypothesis, we examine the distributions of the median router logits throughout training (Fig.~\ref{fig:logit_distrib} for the first layer, App.~\ref{app:logits} for middle and last) for both 1xFLOPs-G8 and 2xFLOPs-G8 models. The plots support the hypothesis: early in training, the router heavily favors the top-1 expert and only gradually learns to utilize additional experts per token over time. This behavior helps to explain why the advantage of fine-grained models becomes more significant with a longer training – the router requires sufficient training data to learn how to effectively distribute load across multiple fine-grained experts.

\vspace{0.4cm}

These findings suggest potential avenues for improving router learning, which could be particularly beneficial for models with even higher granularity. We leave further exploration of MoE router training dynamics for future work.

\vspace{0.5cm}

\subsection{The Order of Softmax and Top-k} \label{sub:soft_topk_order}

The optimal ordering of softmax and Top-k normalization in the MoE router is not clear in the literature. While many works \citep{shazeer2017outrageously, zoph2022stmoedesigningstabletransferable, jiang2024mixtral,vavre2024llama} put the softmax after the Top-$k$ choice, others \citep{fedus2022switch, xue2024openmoeearlyeffortopen, muennighoff2024olmoeopenmixtureofexpertslanguage,he2024upcycling} perform the Top-$k$ choice after softmax. Please note that to preserve gradient in the router, the latter option is necessary if choosing only 1 expert for each input; both are possible when more than 1 expert is selected.

\vspace{0.4cm}

We evaluate the impact of this ordering in our experiments. As presented in Table~\ref{tab:softmax_topk}, for standard MoE models (G1), we observe similar performance between the two approaches (slightly better when applying softmax after Top-k). However, for fine-grained MoE (G8), applying softmax after the Top-k selection yields significantly better results. Consequently, throughout this paper, we use the softmax after Top-k ordering for all models where $k>1$ (1xFLOPS-G8, 2xFLOPS-G8, 2xFLOPS-G1). For the $k=1$ model (1xFLOPS-G1), we necessarily apply softmax before the Top-k selection to maintain gradient flow.

\vspace{0.5cm}

\begin{table}[ht!]
\caption{Validation loss for training models with different orders of softmax and Top-$k$.} 
\label{tab:softmax_topk}
\centering
\resizebox{0.85\textwidth}{!}{
\begin{tabular}{ccc}
\toprule
\multirow{2}{*}{Model} & \multirow{2}{*}{valid loss (softmax before Top-$k$)} & \multirow{2}{*}{valid loss (softmax after Top-$k$)} \\ \\
\midrule
1$\times$FLOPs-G8 & 2.219 & \textbf{2.183} \\
2$\times$FLOPs-G1 & 2.175 & \textbf{2.168} \\
2$\times$FLOPs-G8 & 2.194 & \textbf{2.166} \\
\bottomrule
\end{tabular}
}
\end{table}

\vspace{0.5cm}

One possible explanation of why we observe better performance of softmax after Top-k is that in this case we are guaranteed that all scores for the selected experts sum up to 1. This could support a more stable magnitude of the MoE layer output, positively affecting the training dynamics. We note however, that in practice the difference between the order of performing softmax and Top-k might be reduced by a different router training design or with a long enough training, allowing the router to learn the optimal distribution. 

\pagebreak

\subsection{Summary of Findings for Practitioners}

Below we summarize key findings of this report with practical implications for training MoE models:
\begin{itemize}
\vspace{0.1cm}
    \item \textbf{Performance benefits of fine-grained MoE scale effectively.} Our experiments show that the efficiency advantage of fine-grained MoE architectures scales well for large models (over 50B total parameters) and results in better accuracy on standard benchmarks (Sec. \ref{sec:50b}).
    \vspace{0.1cm}
    \item \textbf{Extended training enhances fine-grained MoE gains.} The performance improvements attributed to fine-grained MoE become more significant as training duration increases (Sec. \ref{sec:longer}).
    \vspace{0.1cm}
    \item \textbf{The order of softmax and Top-$k$ impacts training.} Applying softmax normalization \emph{after}, rather than before, the Top-$k$ selection step yielded better results for fine-grained MoE, underscoring the importance of this router design choice (Sec. \ref{sub:soft_topk_order}).
\end{itemize}

\vspace{0.2cm}

\section{Limitations and Future Work}
In this report, we aim to provide a comprehensive comparison of standard and fine-grained MoE models. However, the scope and focus of this work naturally lead to several avenues for future research.

\vspace{0.1cm}

Notably, a key assumption throughout this report is uniform hardware utilization across different MoE architectures, considering only training steps and FLOPs. In practice, the exact implementation and hardware setup can lead to variations in Model FLOPs Utilization (MFU) depending on the model type. This variability can affect the efficiency gains described here. Implementing efficient training and inference for fine-grained MoE is an important challenge. We refer to related work \citep{tan2024scatteredmixtureofexpertsimplementation,deepseekai2024deepseekv3, gale2022megablocksefficientsparsetraining, doubov2024sparseupcyclinginferenceinefficient,he2024upcycling} and leave a detailed exploration of these implementation and hardware considerations for future studies.

\vspace{0.1cm}

In this work, we employ the standard approach to load balancing \citep{fedus2022switch}, using the auxiliary loss and dropping tokens exceeding the expert capacity. Therefore, while we do not observe expert collapse, other strategies \citep{wang2024auxiliarylossfreeloadbalancingstrategy, deepseekai2024deepseekv3} could improve model quality and hardware utilization.

\vspace{0.1cm}

Another important direction for future work involves further scaling the dataset size. Our experiments present comparisons using setups with a ratio of tokens to active parameters between 6 and 28. This range is close to the $\sim$20 tokens/parameter rule of thumb for dense models as in \citep{hoffmann2022training}. However, many modern LLMs \citep{touvron2023llama2, dubey2024llama3herdmodels, parmar2024nemotron415btechnicalreport} are significantly overtrained to maximize inference efficiency. Therefore, it is important to examine the extent to which our conclusions hold for setups involving trillion-token-scale datasets.

\vspace{0.1cm}

Finally, this work focuses exclusively on the pretraining phase. Unique challenges may emerge during subsequent phases like fine-tuning and post-training, representing another area for future investigation.

\vspace{0.2cm}

\section{Conclusions}

In this work, we empirically evaluated several MoE variants, and proposed training recipes for effectively scaling models up to 56B total (17B active) parameters. Our results show that fine-grained MoE improves model quality and computational efficiency compared to the standard architecture. 
Specifically, we found that using higher granularity leads to lower validation loss and better downstream benchmark scores. The benefit was particularly noticeable with longer training. 
These findings provide empirical and practical grounding for training future large-scale fine-grained MoE models.

\clearpage

\section*{Acknowledgments}
We thank Ethan He for his valuable comments and suggestions throughout this project. We are grateful to Janusz Lisiecki and to Jacek Staszewski and Małgorzata Ciechomska for their instrumental support in establishing the collaboration between NVIDIA and IDEAS NCBR. We also extend our thanks to Maciej Pióro, Sebastian Jaszczur and Jan Ludziejewski for their helpful feedback on this report. Furthermore, we wish to express the gratitude to Piotr Sankowski for the creation of the supportive scientific environment.

\printbibliography

\clearpage
\appendix
\section{Distribution of the Router Logits}\label{app:logits}

\begin{figure}[ht!]
    \centering
    \subfigure{
        \includegraphics[width=0.17\textwidth]{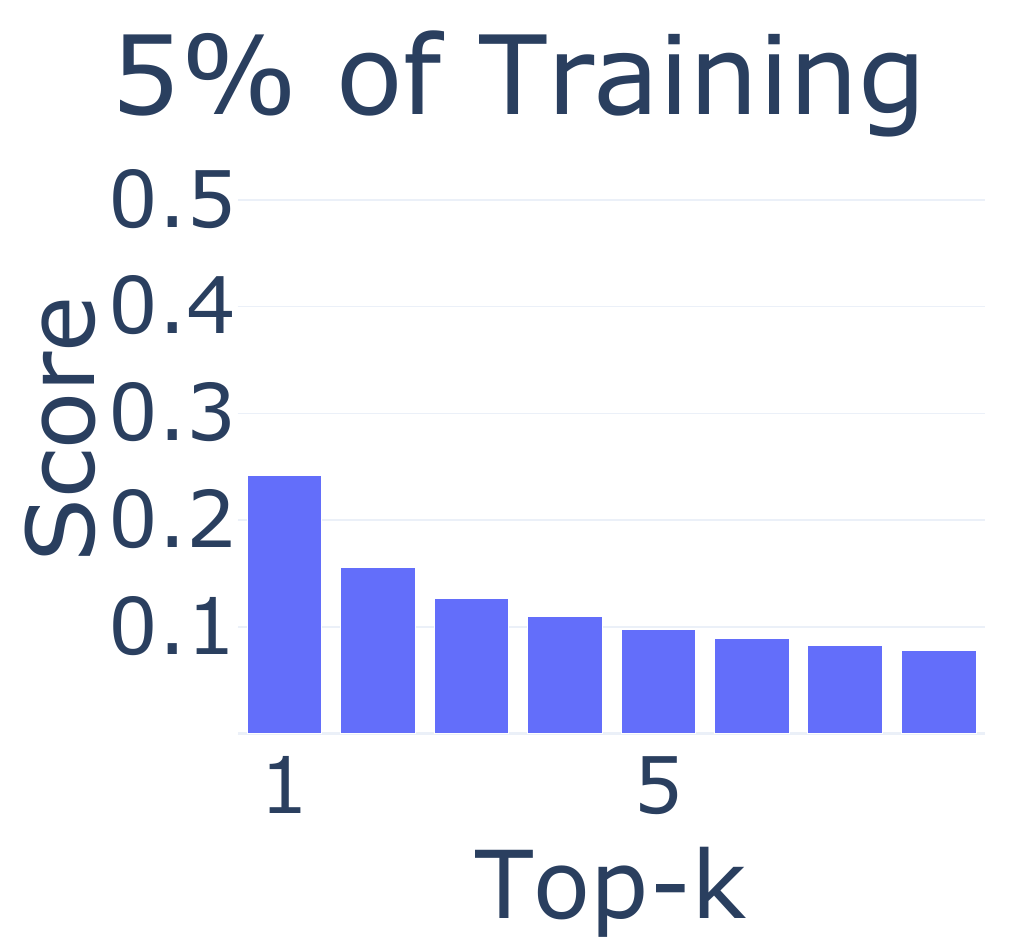}
    }
    \subfigure{
        \includegraphics[width=0.17\textwidth]{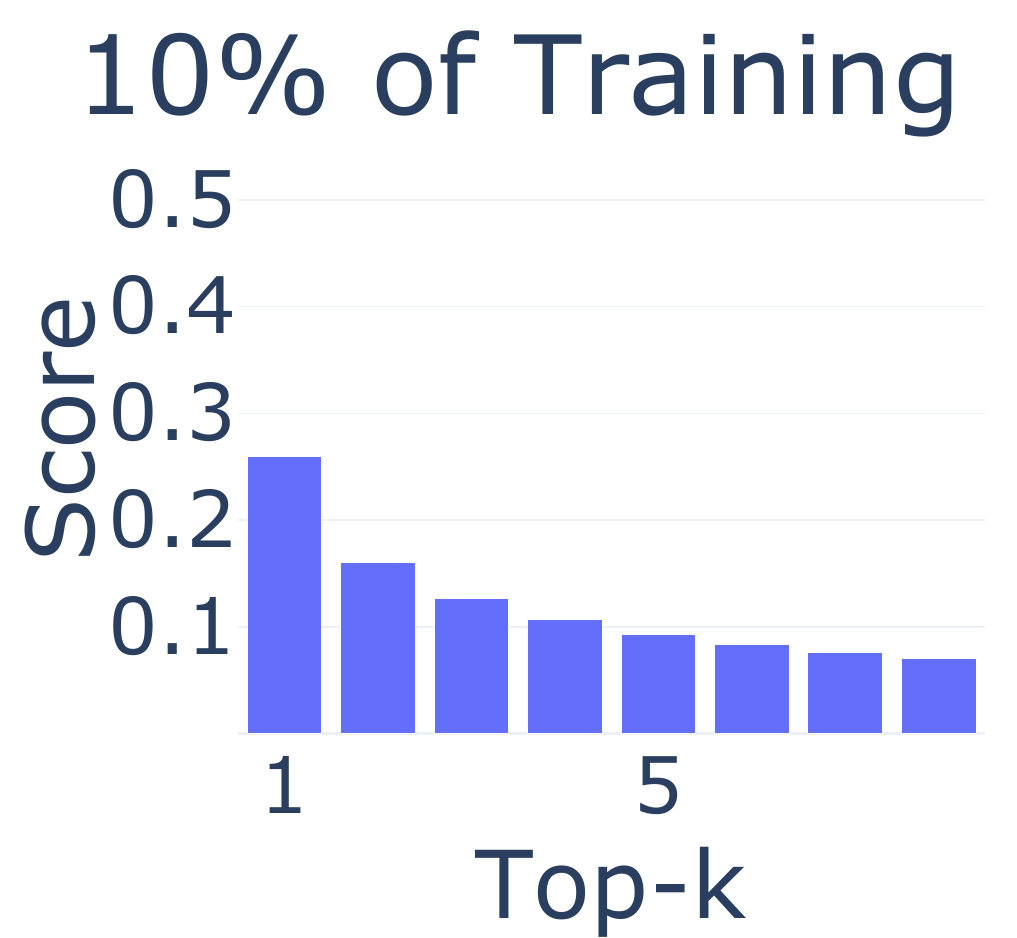}
    }
    \subfigure{
        \includegraphics[width=0.17\textwidth]{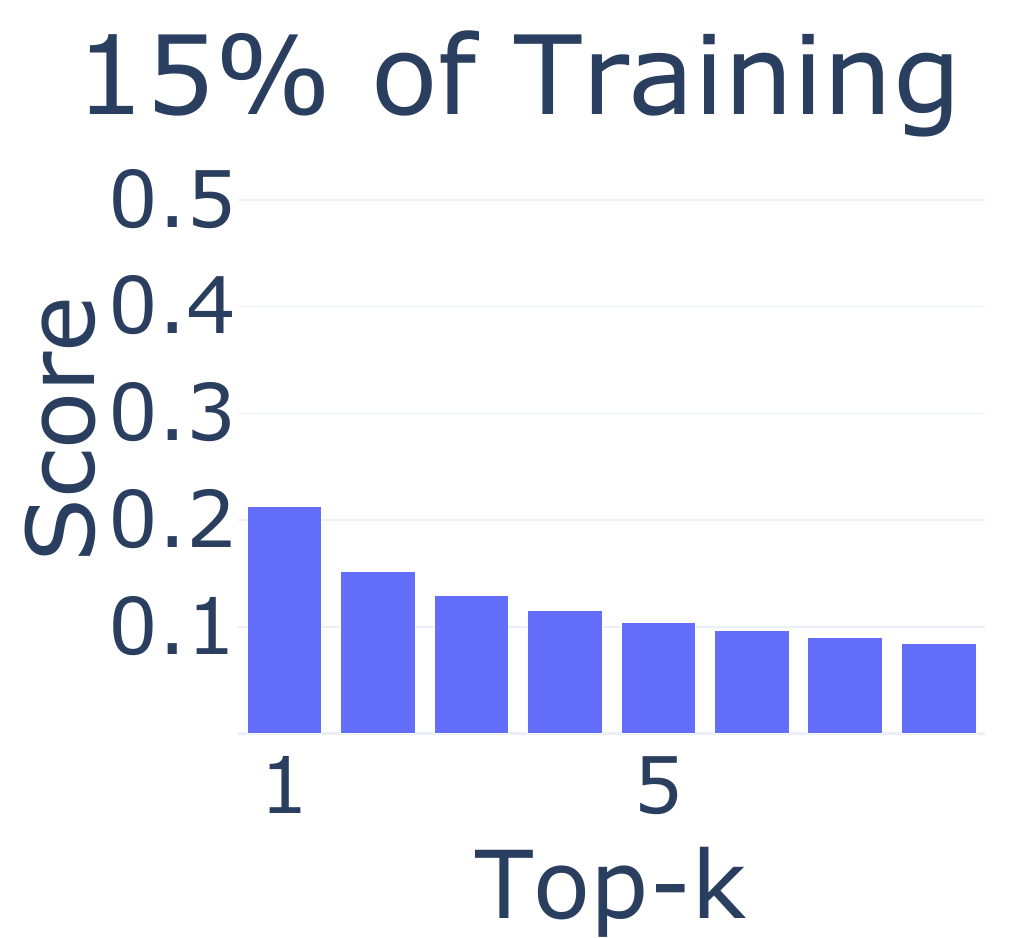}
    }
    \subfigure{
        \includegraphics[width=0.17\textwidth]{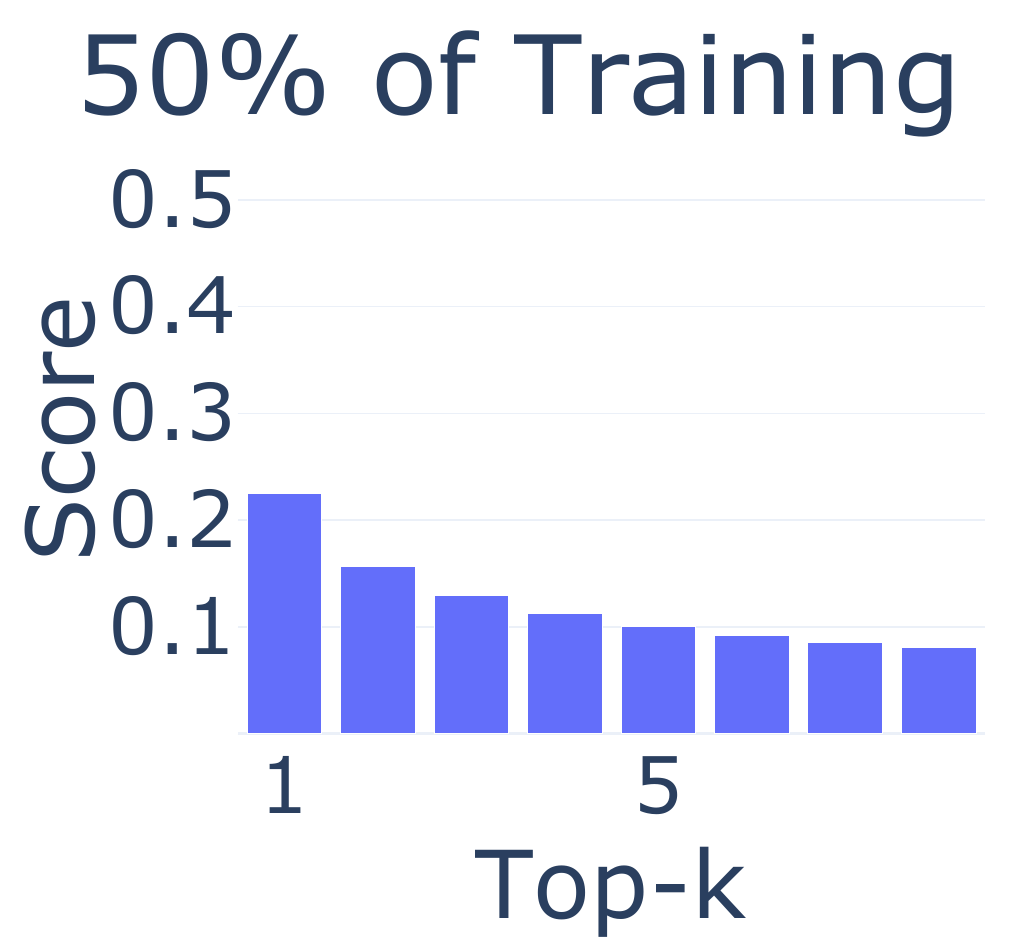}
    }
    \subfigure{
        \includegraphics[width=0.17\textwidth]{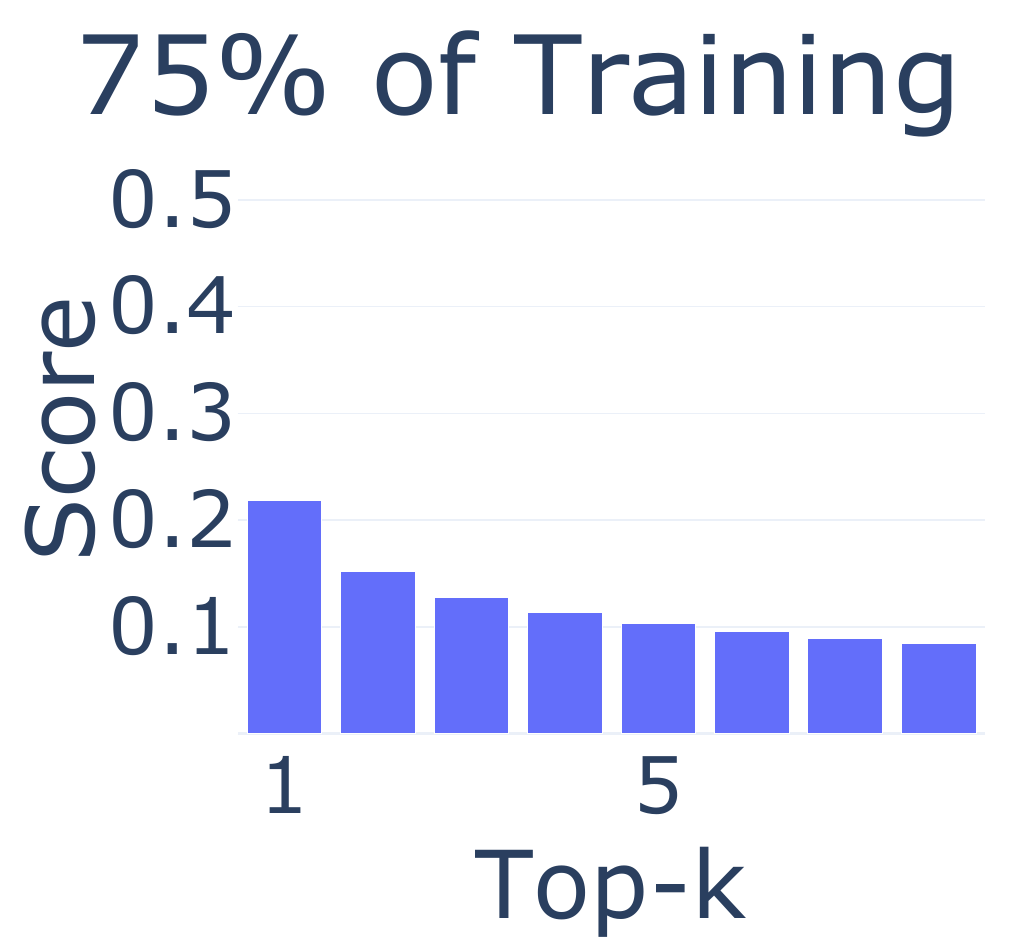}
    }\\
\small\textit{top}: 1xFLOPs-G8, middle layer (layer 12)

\par\vspace{0.2em} 

    \subfigure{
        \includegraphics[width=0.17\textwidth]{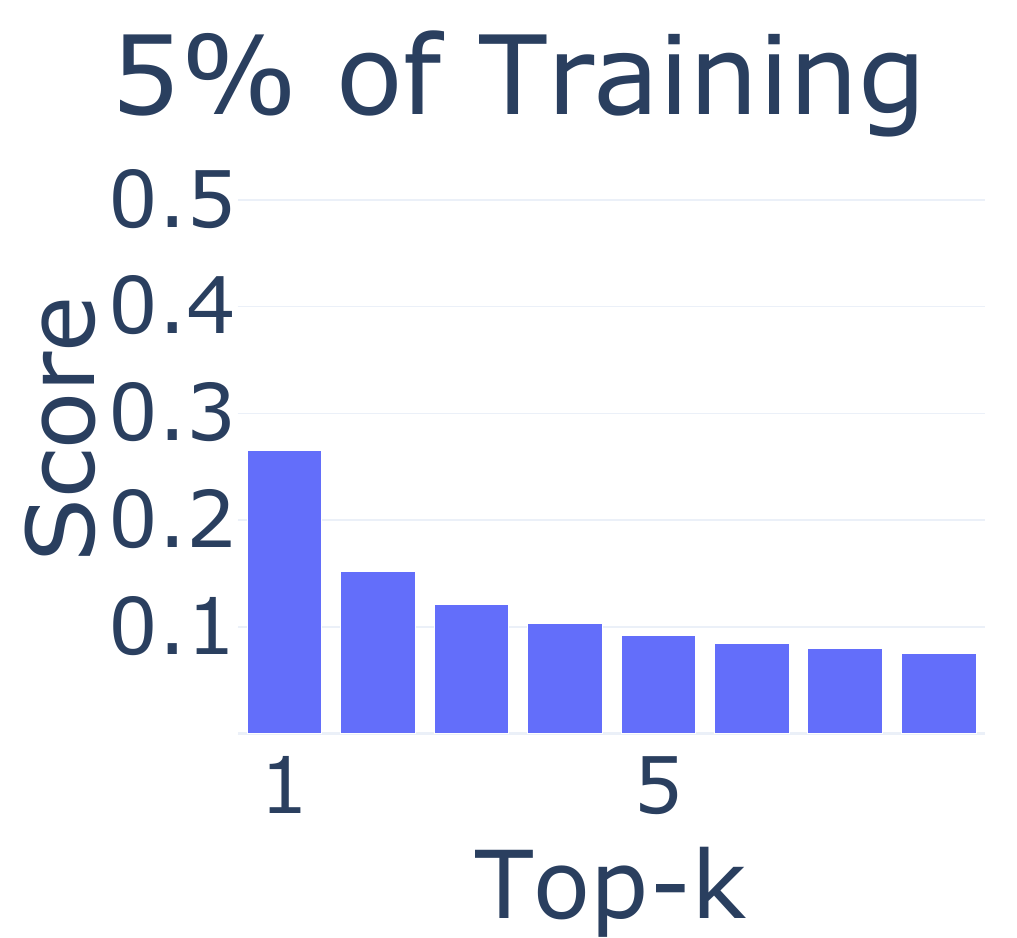}
    }
    \subfigure{
        \includegraphics[width=0.17\textwidth]{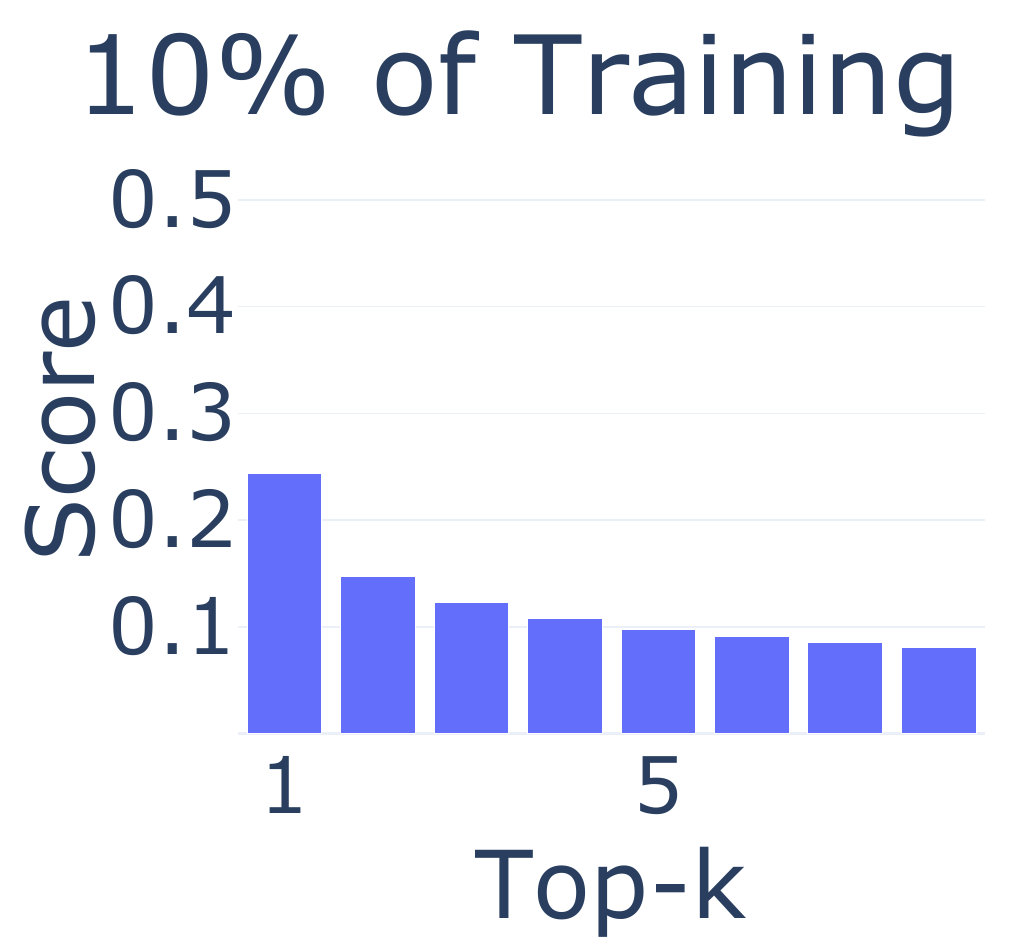}
    }
    \subfigure{
        \includegraphics[width=0.17\textwidth]{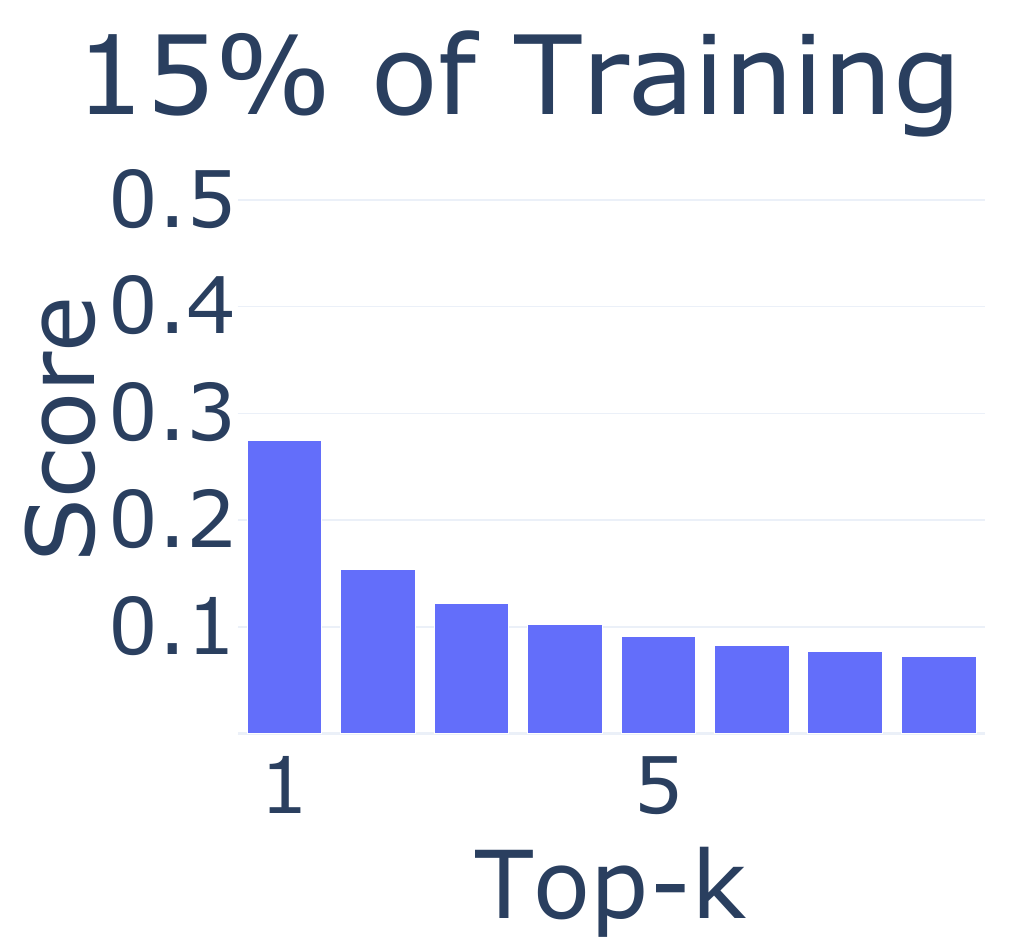}
    }
    \subfigure{
        \includegraphics[width=0.17\textwidth]{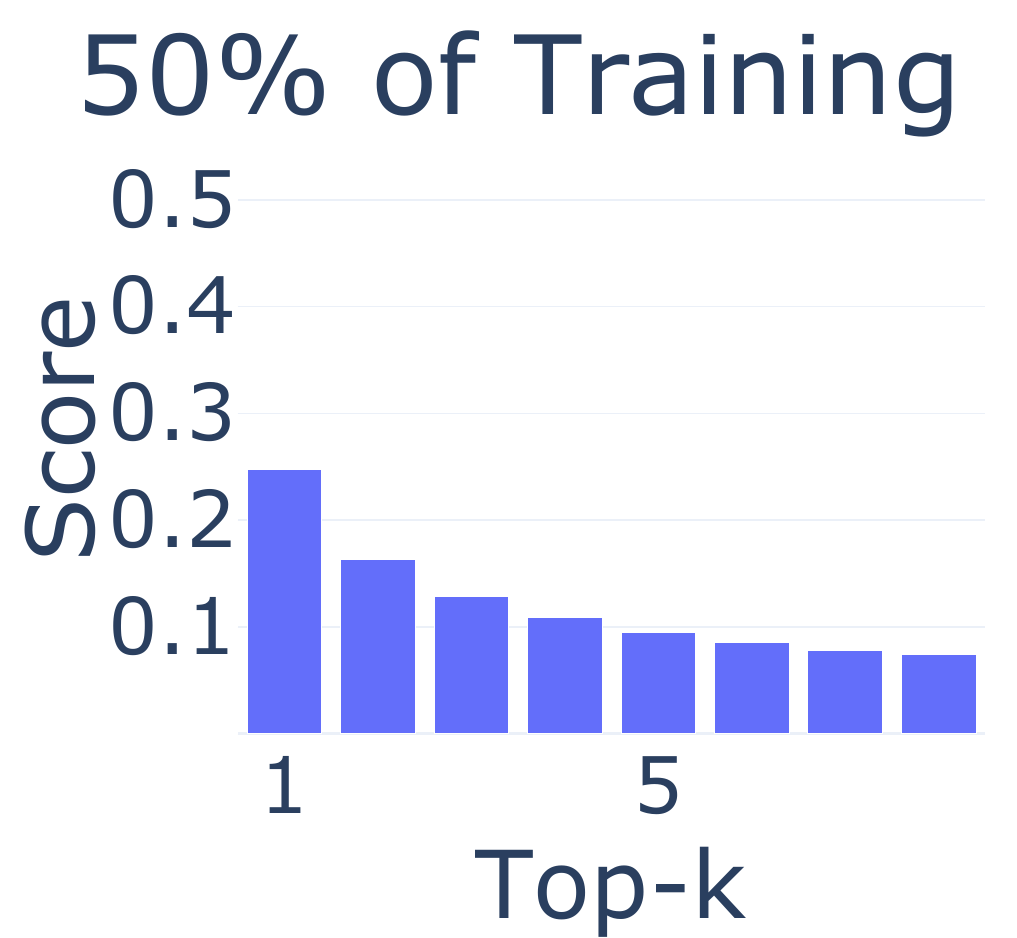}
    }
    \subfigure{
        \includegraphics[width=0.17\textwidth]{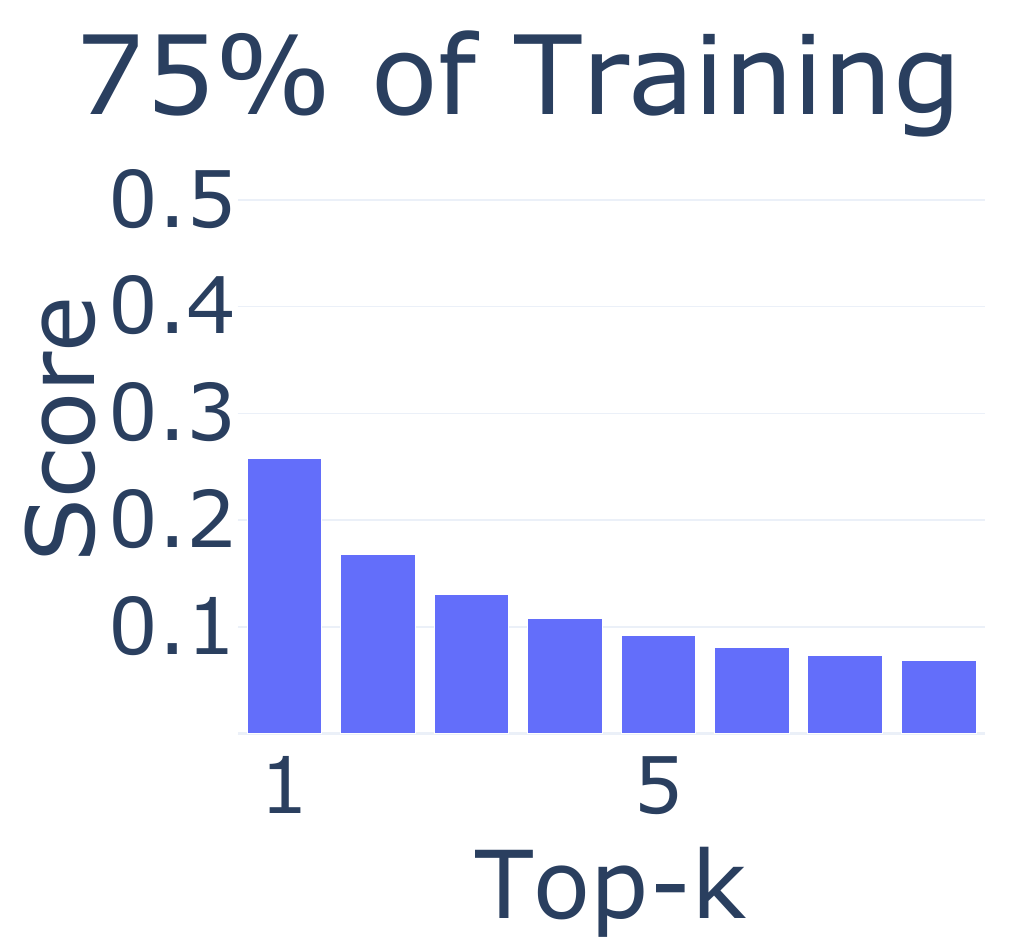}
    }\\
\small\textit{bottom}: 1xFLOPs-G8, final layer (layer 24)
\caption{Distribution of router logits in the middle \textit{(top)} and final \textit{(bottom)} layer of the 1xFLOPs-G8 model.}
    
  \label{fig:logit_distrib_app}
\end{figure}

\begin{figure}[ht!]
    \centering
    \subfigure{
        \includegraphics[width=0.17\textwidth]{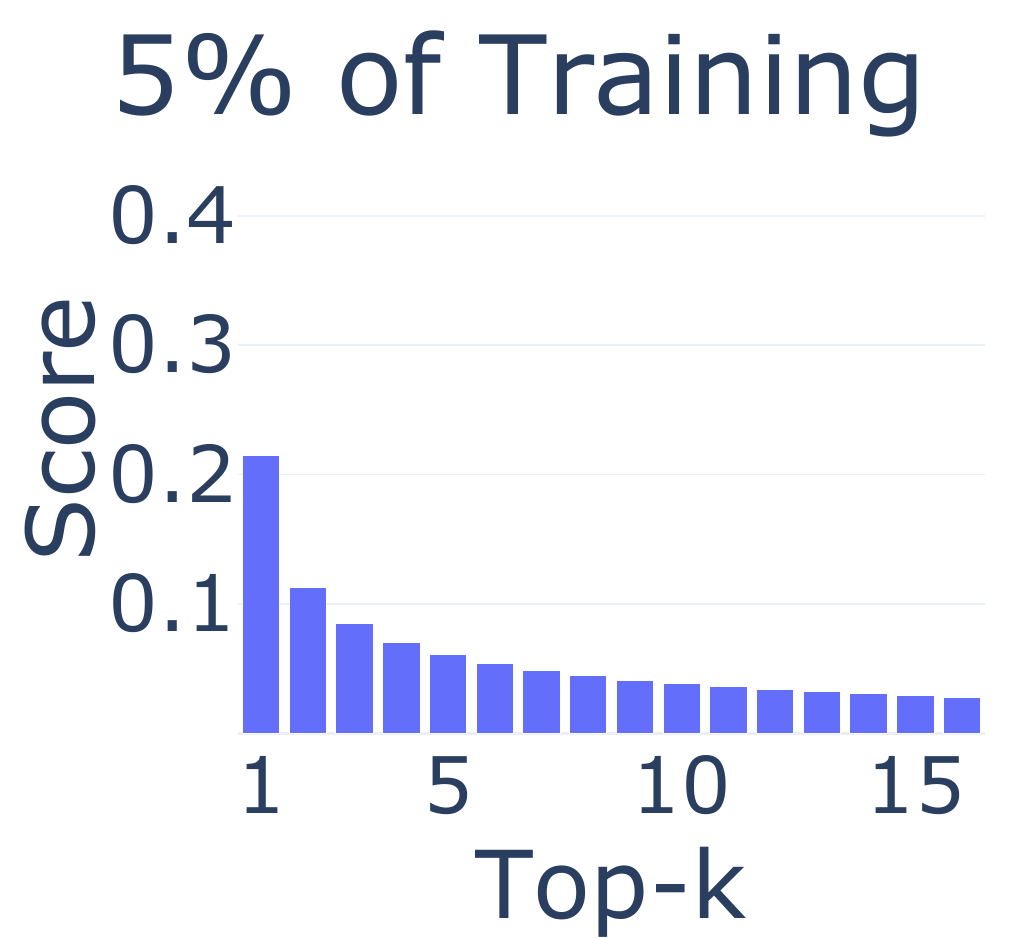}
    }
    \subfigure{
        \includegraphics[width=0.17\textwidth]{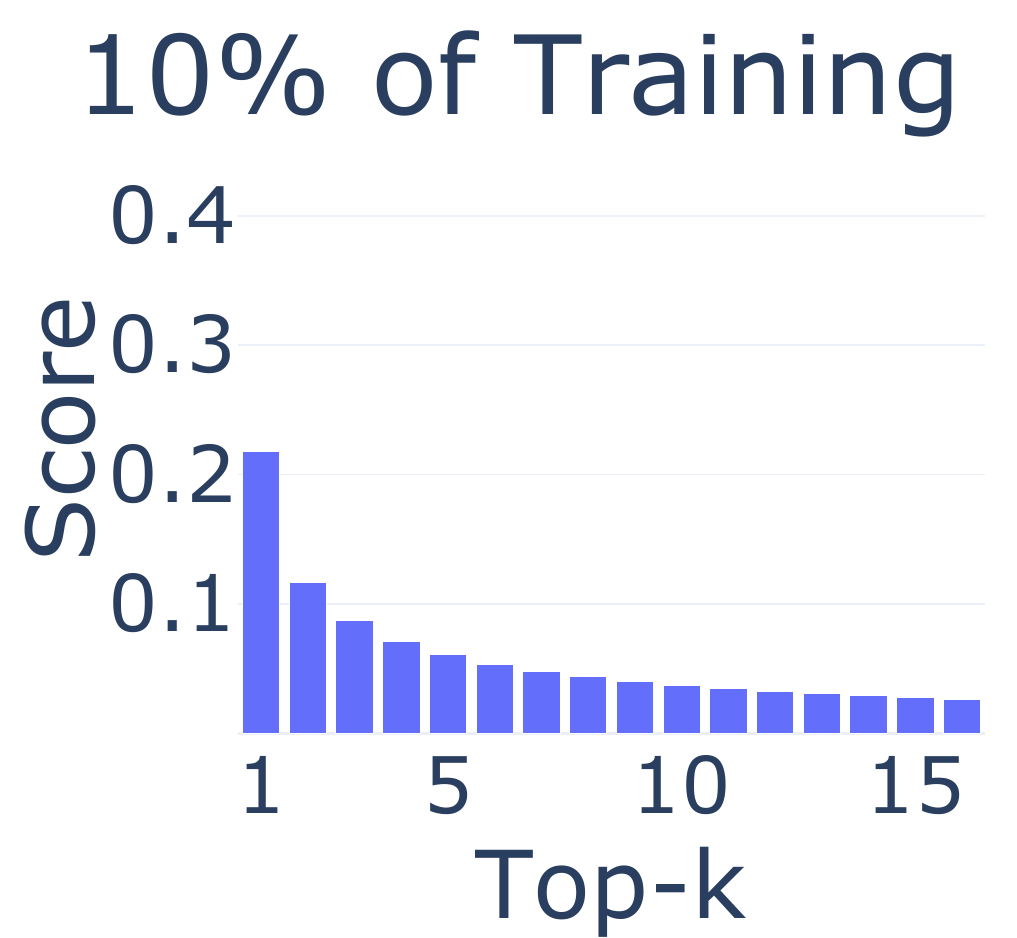}
    }
    \subfigure{
        \includegraphics[width=0.17\textwidth]{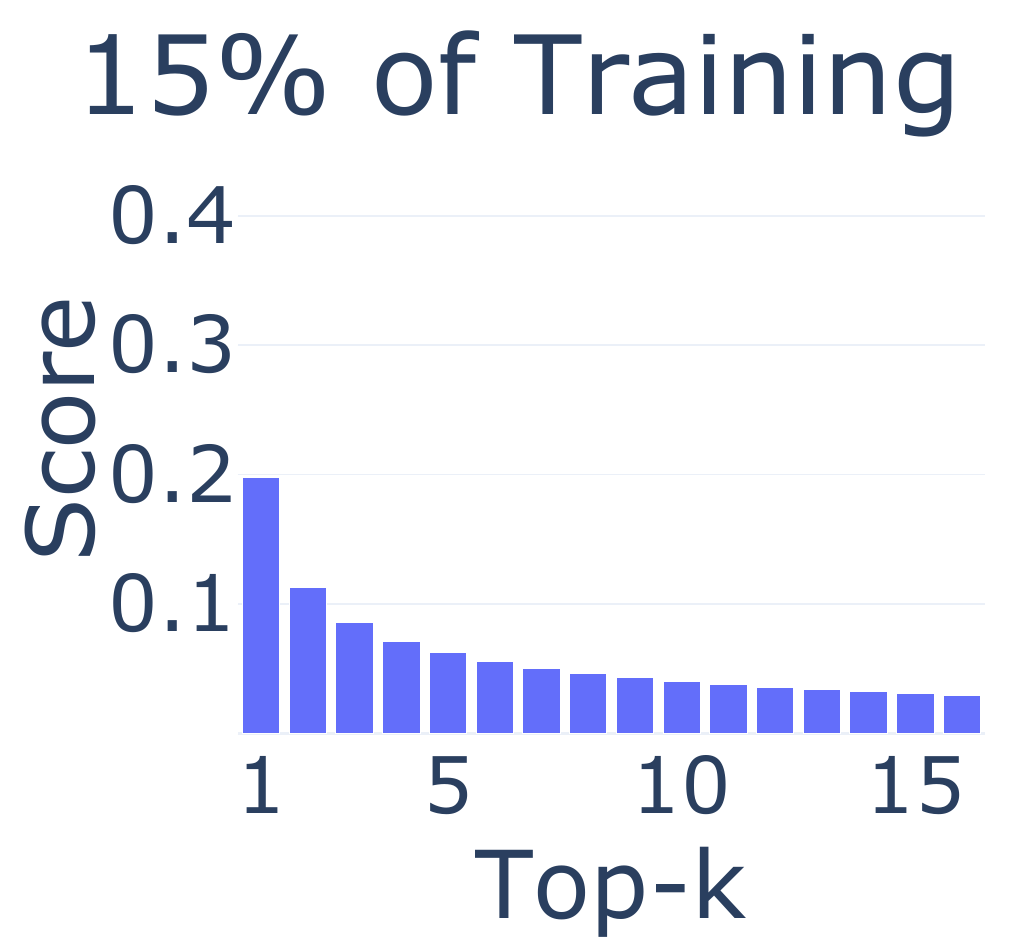}
    }
    \subfigure{
        \includegraphics[width=0.17\textwidth]{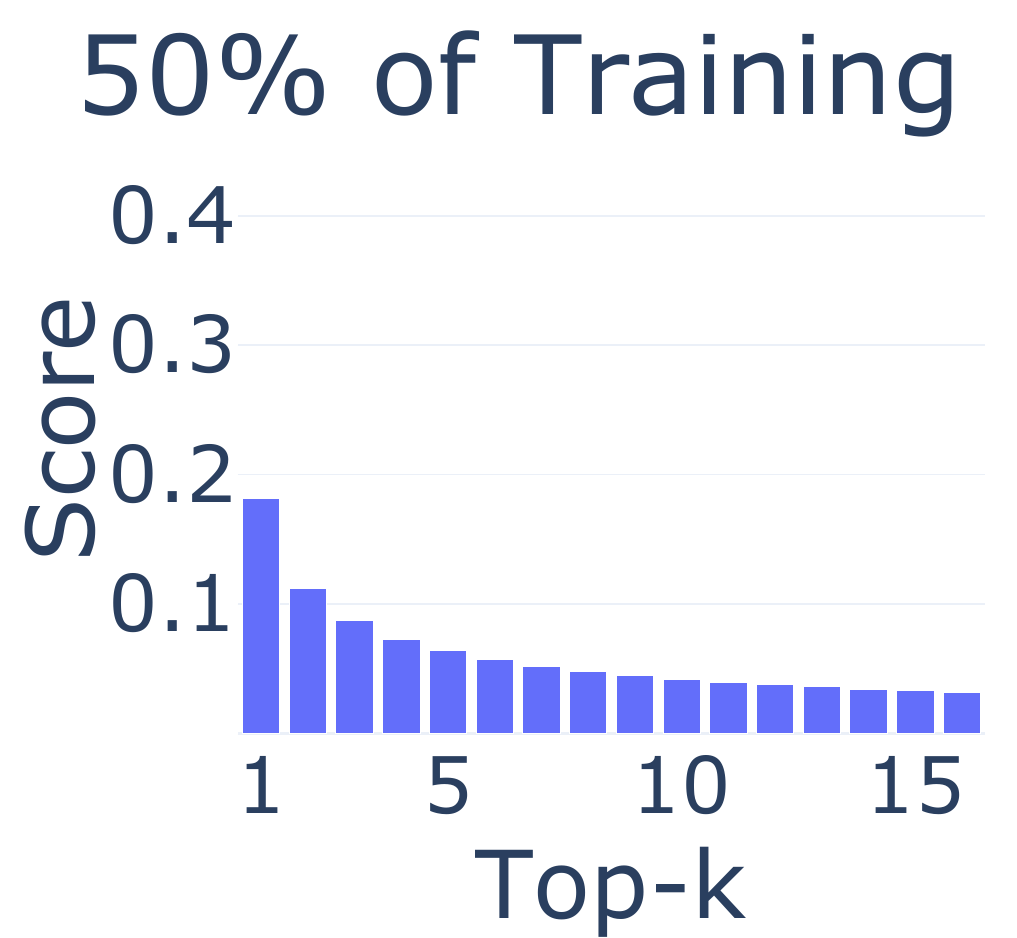}
    }
    \subfigure{
        \includegraphics[width=0.17\textwidth]{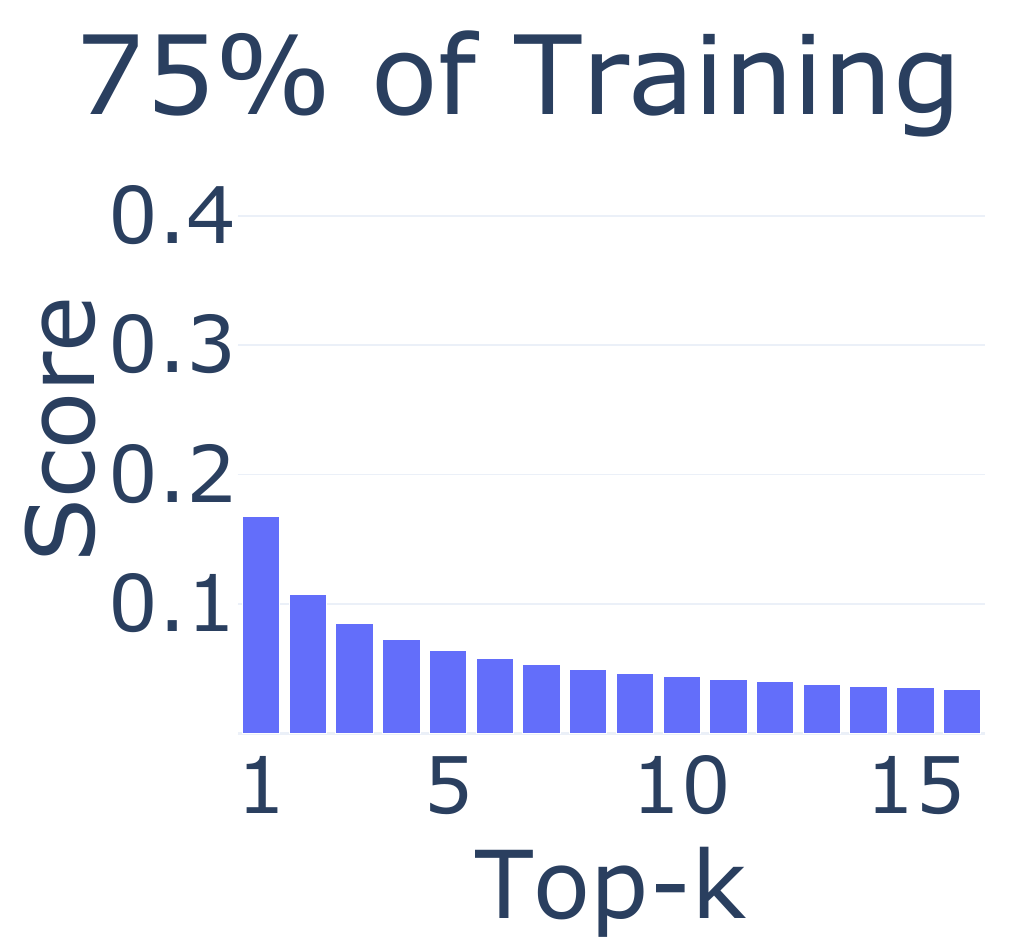}
    }\\
\small\textit{top}: 2xFLOPs-G8, middle layer (layer 12)

\par\vspace{0.2em} 

    \subfigure{
        \includegraphics[width=0.17\textwidth]{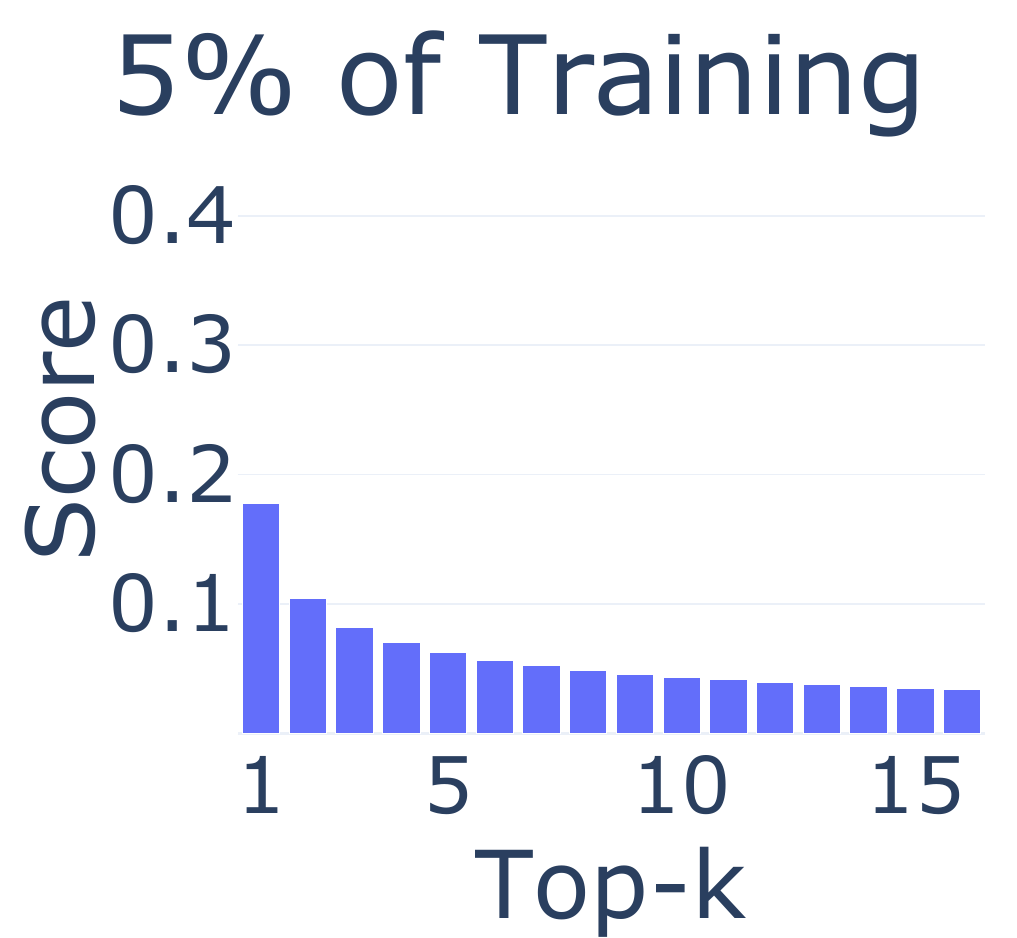}
    }
    \subfigure{
        \includegraphics[width=0.17\textwidth]{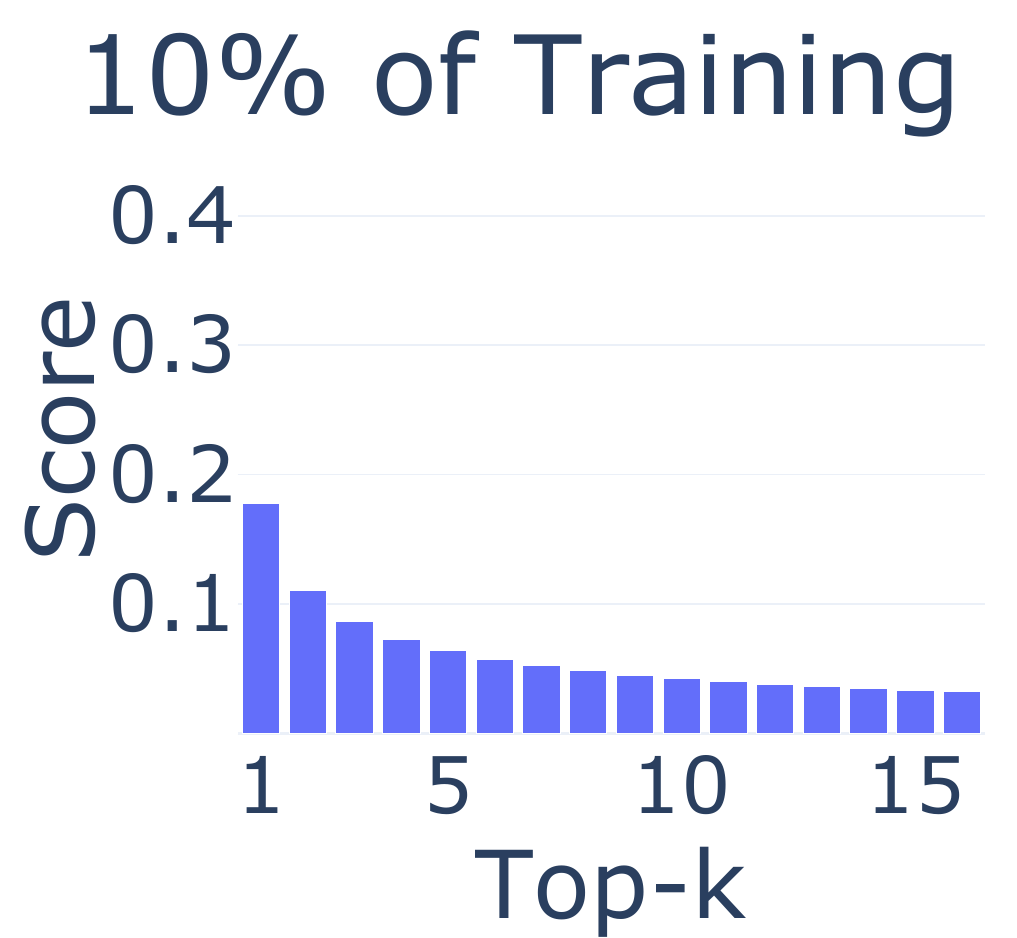}
    }
    \subfigure{
        \includegraphics[width=0.17\textwidth]{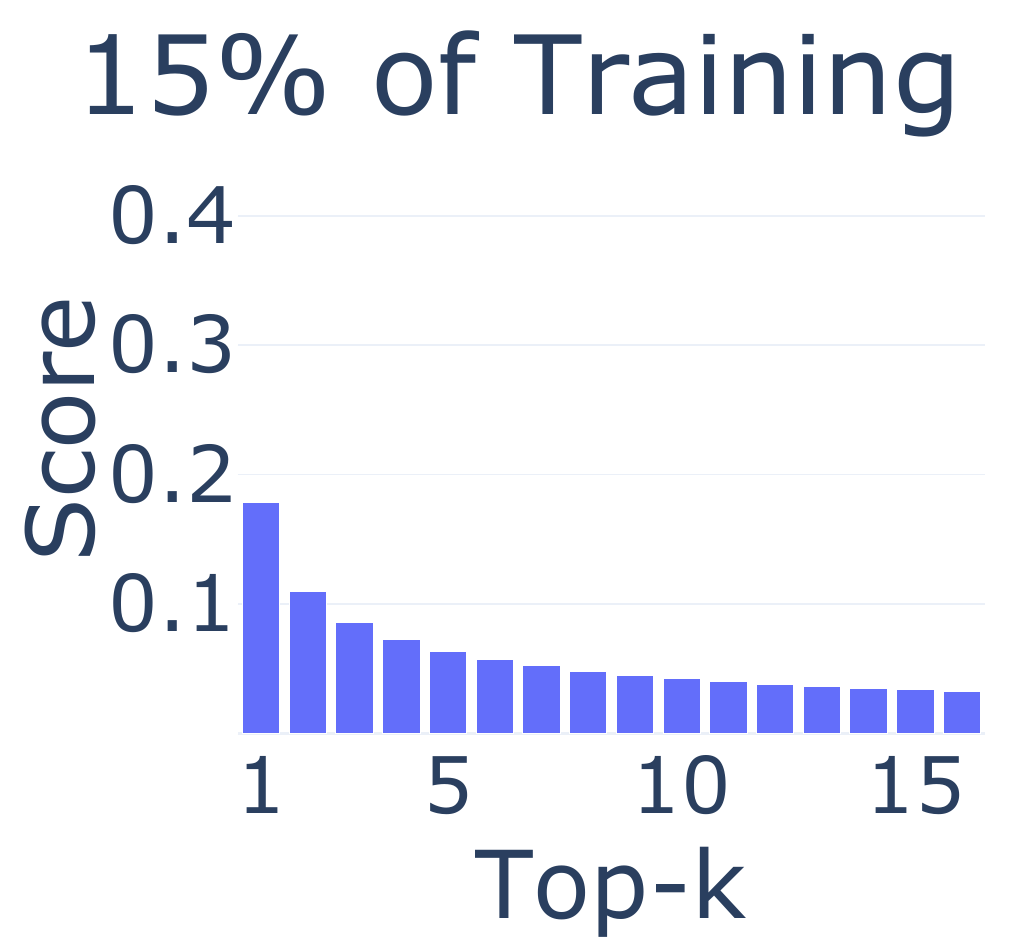}
    }
    \subfigure{
        \includegraphics[width=0.17\textwidth]{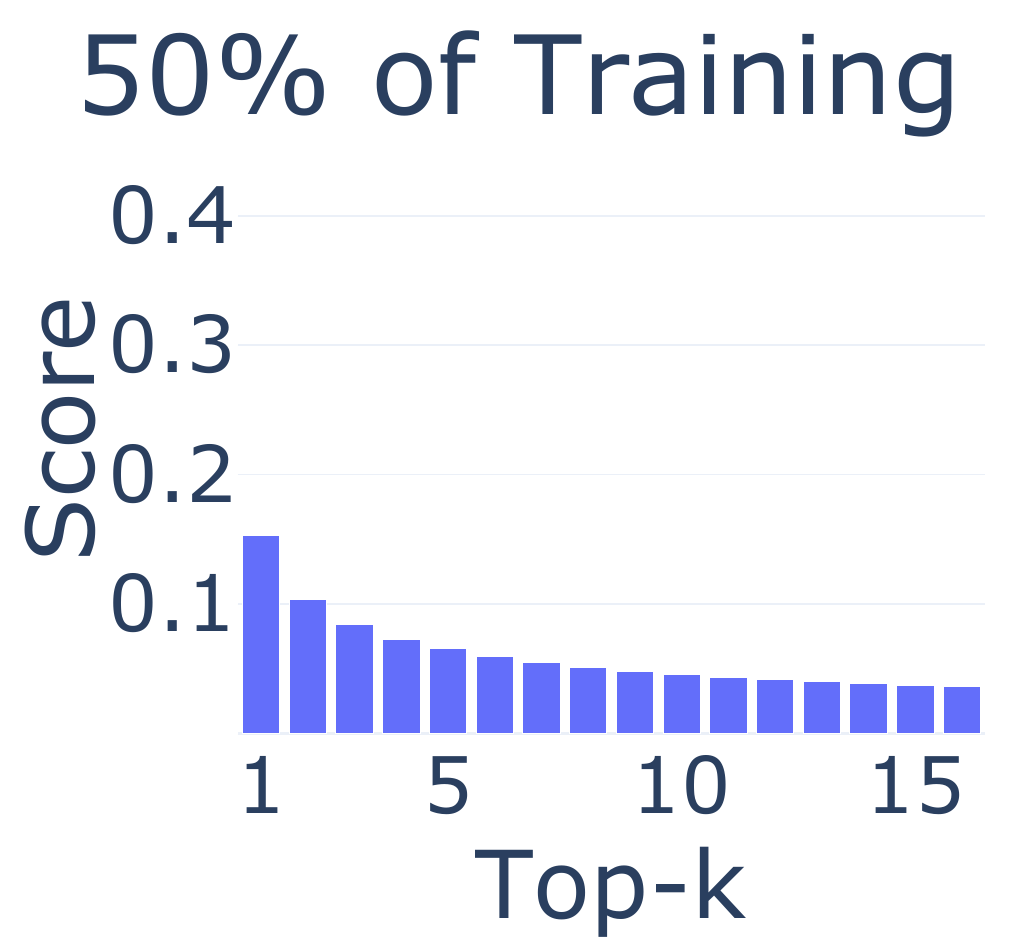}
    }
    \subfigure{
        \includegraphics[width=0.17\textwidth]{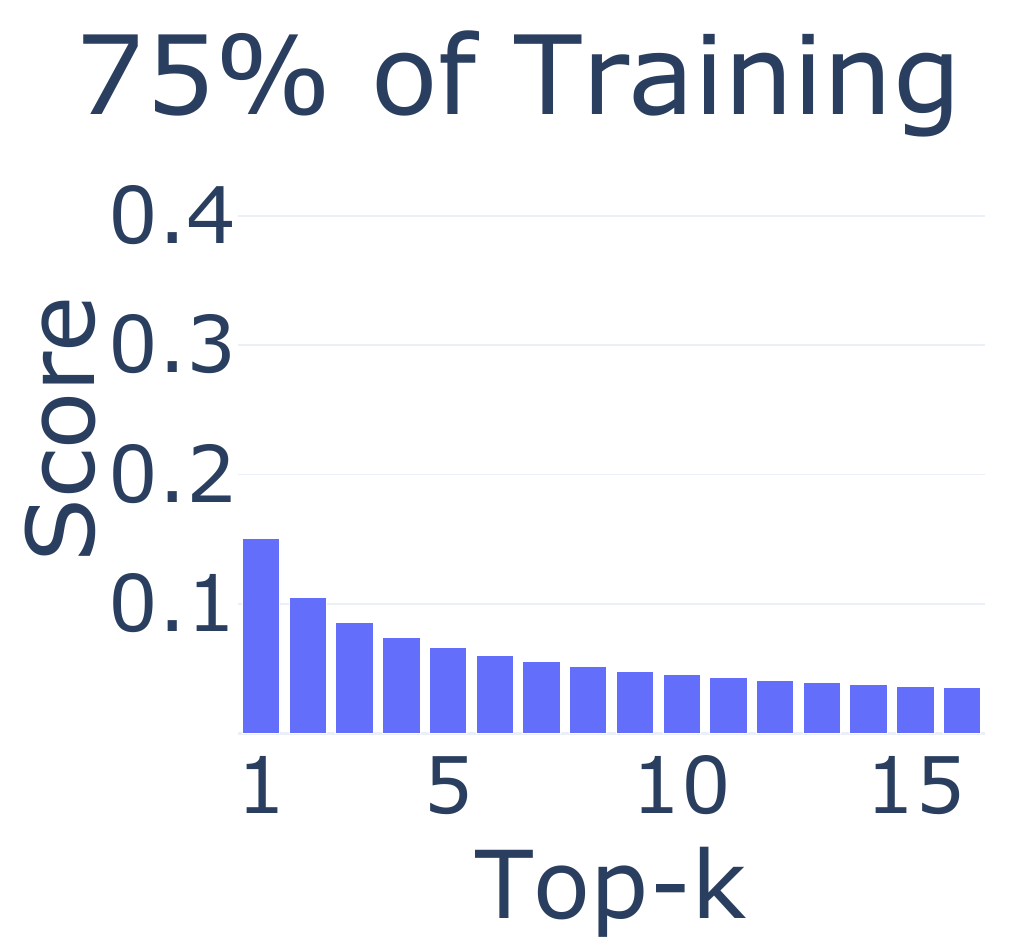}
    }\\
\small\textit{bottom}: 2xFLOPs-G8, final layer (layer 24)
\caption{Distribution of router logits in the middle \textit{(top)} and final \textit{(bottom)} layer of the 2xFLOPs-G8 model.}
    
  \label{fig:logit_distrib_app_2}
\end{figure}


\clearpage

\section{Analysis of Load Balance} \label{app:imbalance}

\begin{figure}[ht!]
    \centering
    \subfigure{
        \includegraphics[width=0.48\textwidth]{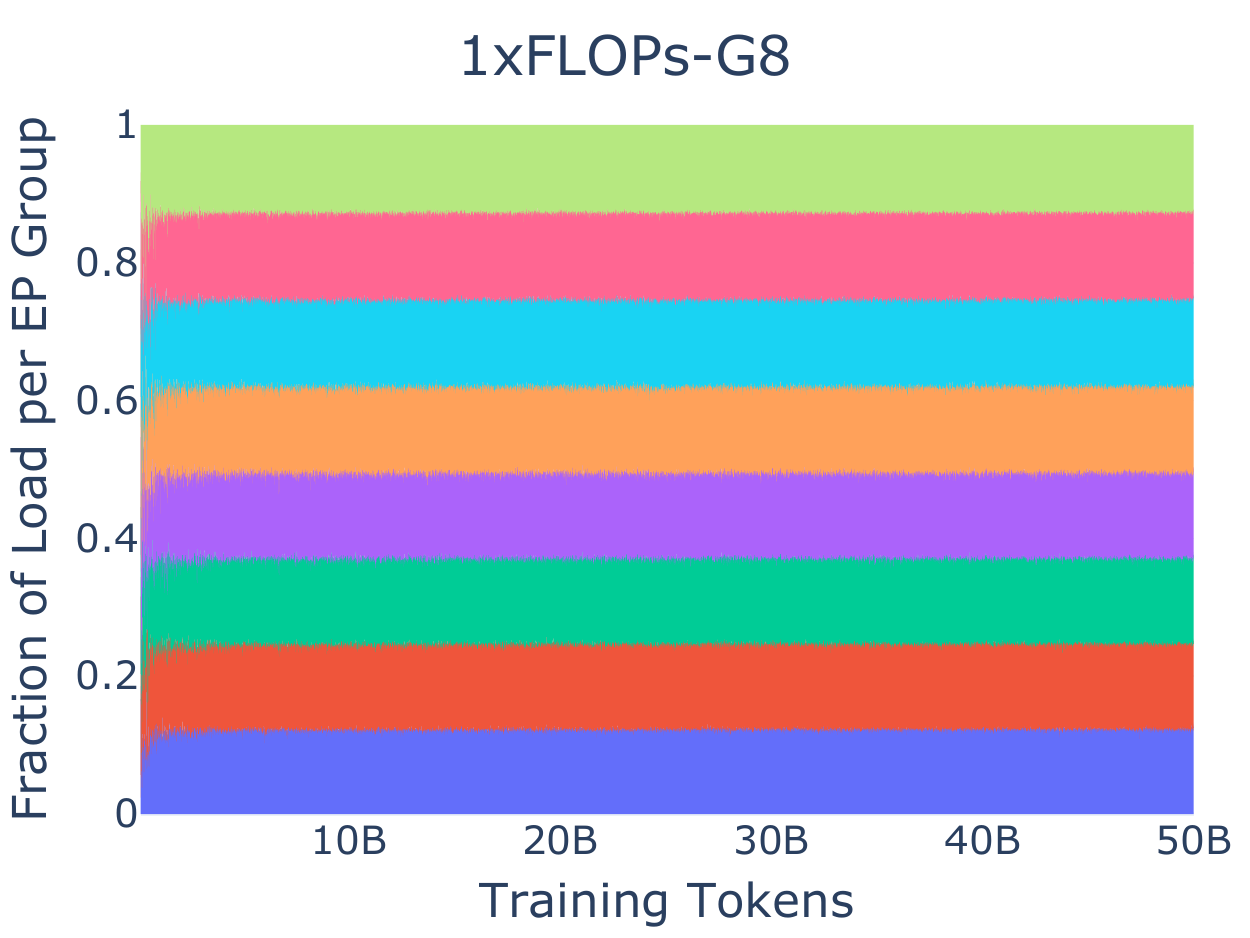}
    }
    \subfigure{
        \includegraphics[width=0.48\textwidth]{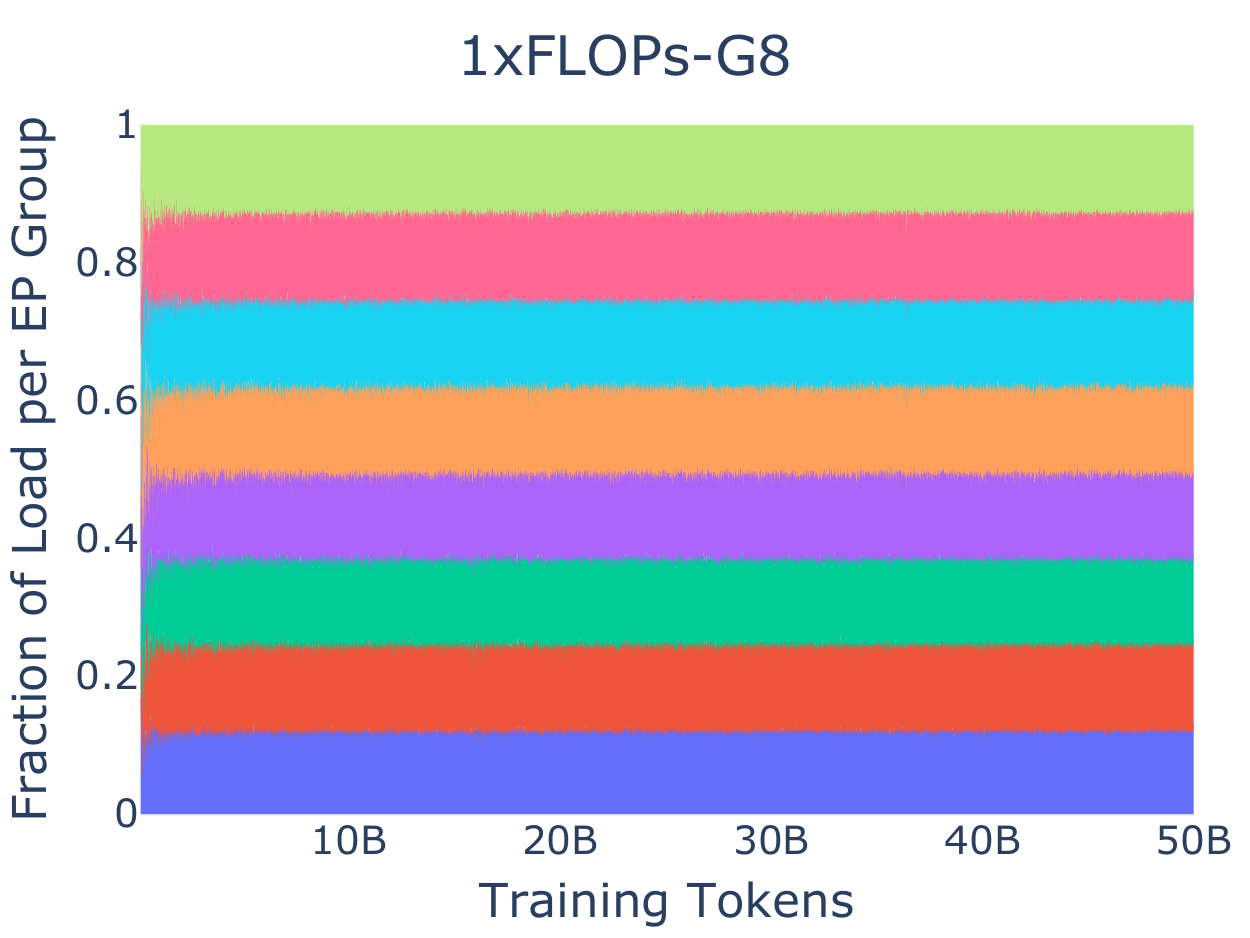}
    }
\caption{Fraction of load assigned to each Expert Parallel group for the 11B 1xFLOPs-G8 model. \textit{(left)}: Distribution in the middle (12th) layer. \textit{(right)}: Distribution in the final (24th) layer.}
    
  \label{fig:ep_load_app}
\end{figure}

\begin{figure}[ht!]
    \centering
    \subfigure{
        \includegraphics[width=0.48\textwidth]{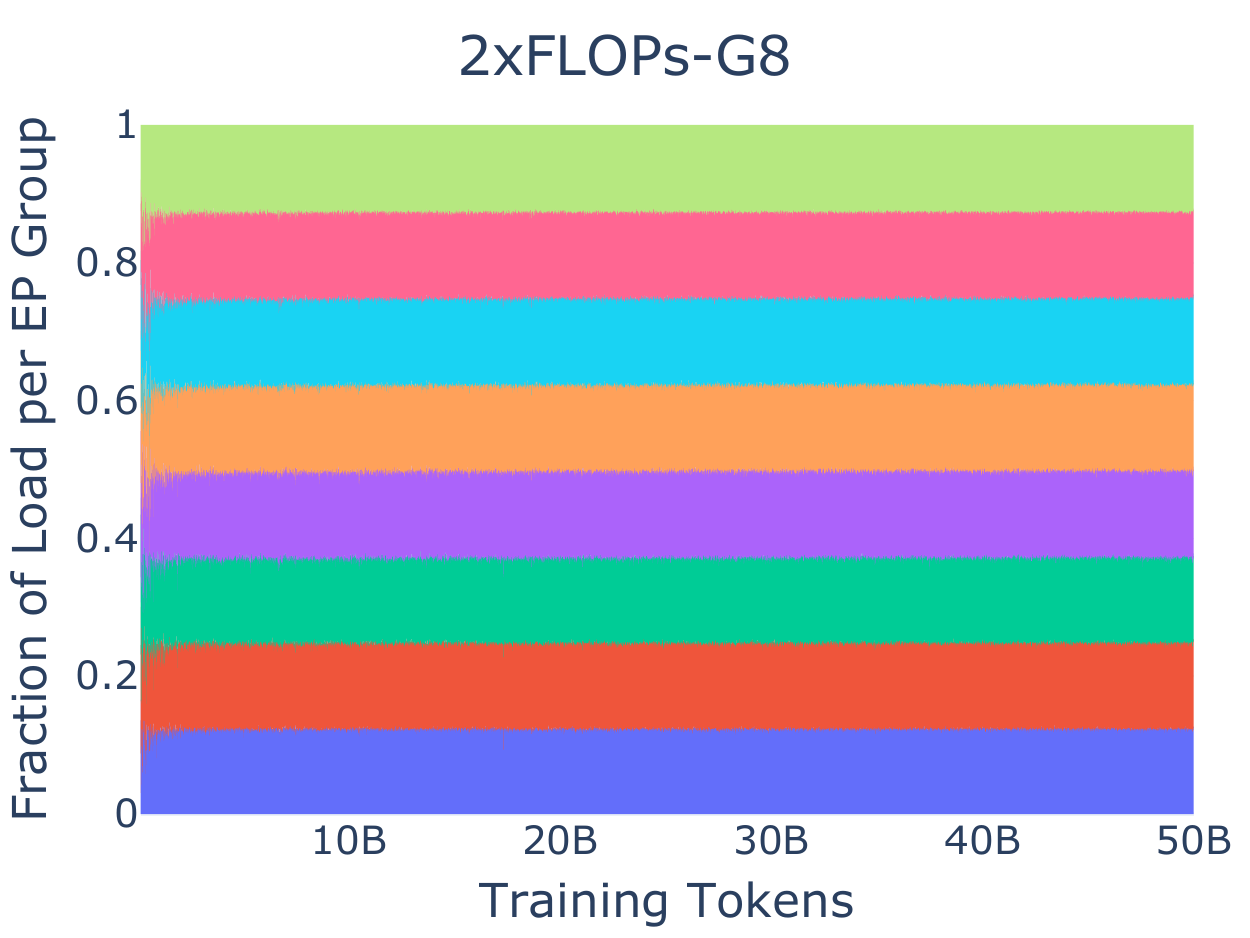}
    }
    \subfigure{
        \includegraphics[width=0.48\textwidth]{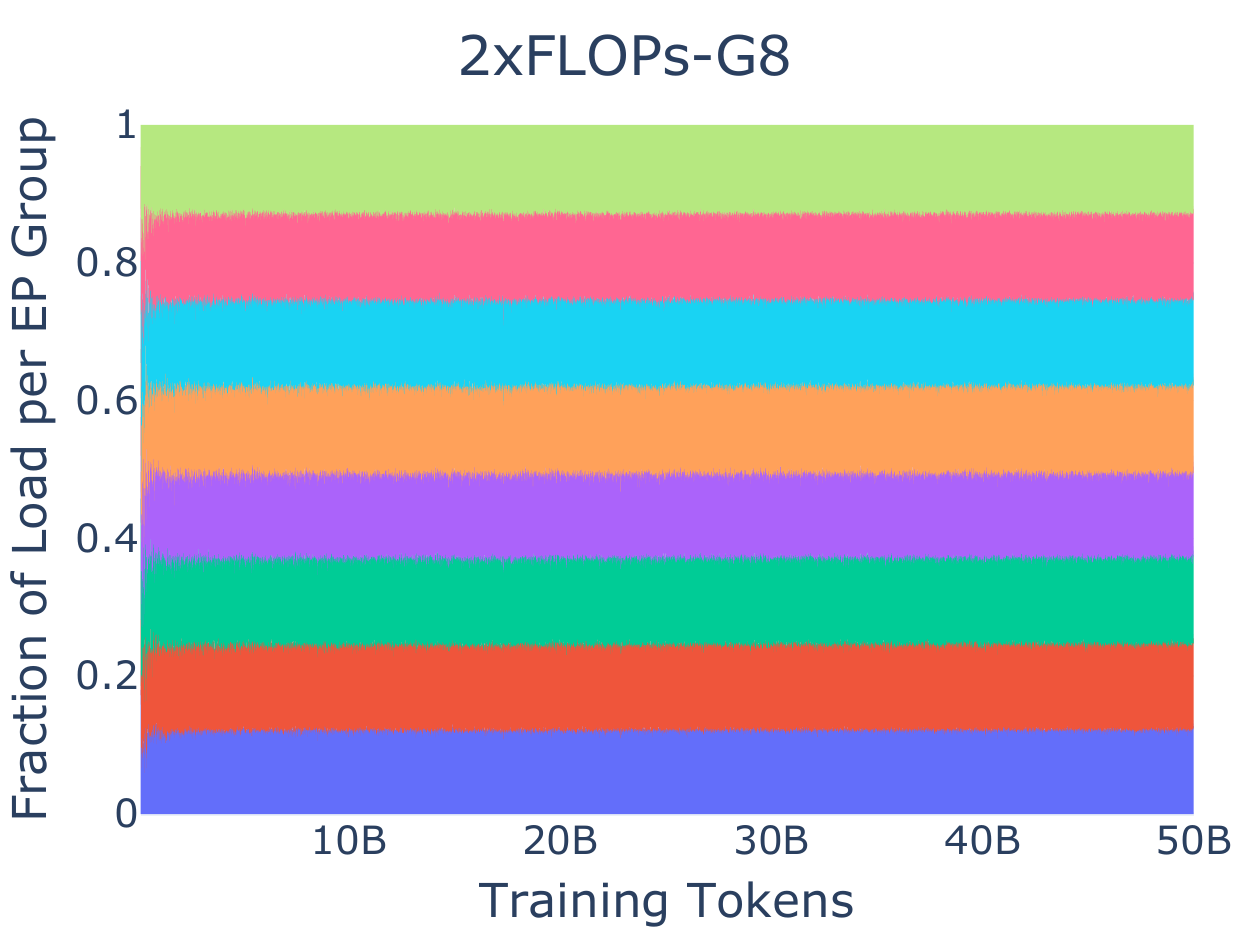}
    }
\caption{Fraction of load assigned to each Expert Parallel group for the 11B 2xFLOPs-G8 model. \textit{(left)}: Distribution in the middle (12th) layer. \textit{(right)}: Distribution in the final (24th) layer.}
    
  \label{fig:ep_load_app_2}
\end{figure}

\end{document}